%% file: allinone.tex
\documentclass[review,anon]{llncs}
\usepackage{graphicx,subfigure}
\usepackage{epsfig}
\usepackage{hyperref}
\usepackage{amssymb}
\usepackage{wrapfig}
\usepackage[utf8]{inputenc}
\usepackage{listings}
\usepackage{{booktabs}}
\usepackage{pdfpages}
\usepackage{svg}
\usepackage[ruled,vlined,linesnumbered,titlenotnumbered,noend]{algorithm2e}

\SetCommentSty{mycommfont}

\usepackage{scalerel}
\usepackage{enumitem}
\usepackage{xspace}
\usepackage{rotating}

\usepackage{empheq}
\usepackage{verbatim}
\usepackage{cleveref}
\usepackage{amsmath}
\usepackage{graphicx}
\usepackage{color}

\usepackage{makecell}
\usepackage{array}

\usepackage{lineno}
\usepackage[normalem]{ulem}

\definecolor{cadmiumgreen}{rgb}{0.0, 0.42, 0.24}
\definecolor{verde}{rgb}{0.25,0.5,0.35}                                                 \definecolor{jpurple}{rgb}{0.5,0,0.35}                                                  \definecolor{darkgreen}{rgb}{0.0, 0.2, 0.13}                                            

\usepackage{tikz}
\usetikzlibrary{automata,positioning,decorations.markings,arrows,intersections, backgrounds}
\usetikzlibrary{arrows, positioning, automata}

\colorlet{darkgreen}{green!40!black}
\colorlet{darkblue}{blue!60!black}
\colorlet{darkred}{red!50!black}
\colorlet{safecellcolor}{yellow!5}
\colorlet{goodcellcolor}{green!10}
\colorlet{badcellcolor}{blue!10}

\tikzset{
  >=latex,node distance=2cm,on grid,auto, initial text=,
  box state/.style={draw,rectangle,minimum size=8mm,rounded corners},
  prob state/.style={draw,very thick,shape=circle,darkblue,minimum size=3mm,inner sep=0mm},
  every loop/.style={shorten >=0pt},
  accepting state/.style={double distance=1.2pt, outer sep = 0.6pt+\pgflinewidth},
  accepting dot/.style={above=-2.7pt,circle,fill,darkgreen,inner sep=2pt,radius=1pt},
  loop above/.append style={every loop/.append style={out=120, in=60, looseness=6}},
  loop below/.append style={every loop/.append style={out=300, in=240, looseness=6}},
  loop left/.append style={every loop/.append style={out=210, in=150, looseness=6}},
  loop right/.append style={every loop/.append style={out=30, in=330, looseness=6}}
}

\usetikzlibrary{positioning, shapes}

\DeclareTextFontCommand{\textsfsl}{\sffamily\slshape}
\newcommand{\ourtool}{\textsf{Janaka }\textnormal{}}
\newcommand{\semanticsone}{\textsf{Robust Semantics }\textnormal{}}
\newcommand{\semanticstwo}{\textsf{Discounted Semantics }\textnormal{}}
\newcommand{\chatgpt}{\mathsf{ChatGPT 4.0}}
\newcommand{\gurobi}{\mathsf{Gurobi}}
\newcommand{\bard}{\mathsf{Bard}}
\newcommand{\claude}{\mathsf{Claude}}

\DeclareMathOperator*{\argmax}{argmax}

\newcommand{\sem}[1]{ [ \! [ {#1}  ]  \! ]} % semantic brackets
\newcommand{\set}[1]{\left\{ #1 \right\}}

\newcommand{\seq}[1]{\langle #1 \rangle}

\usepackage{temporal}

\usepackage{stmaryrd}

\newcommand{\limplies}{\Rightarrow}

\newcommand{\union}{\cup}

\newcommand*{\sample}{\mathcal{S}}

\newcommand*{\op}{\mathcal{O}}
\newcommand*{\prop}{\mathcal{P}}

\newcommand*{\finally}{\textsf{F}}
\newcommand*{\globally}{\textsf{G}}
\newcommand*{\Next}{\textsf{X}}

\newcommand{\dom}{\mathsf{dom}}
\newcommand{\depth}{\ensuremath{d}}
\newcommand{\nodes}{\mathcal{N}}

\newcommand{\arrr}{\alpha}

\begin{document}

\pagenumbering{arabic}

\title{Integrating Explanations in Learning LTL Specifications from Demonstrations}
\author {Ashutosh Gupta\inst{2} \and
        John Komp\inst{1} \and
        Abhay Singh Rajput\inst{2} \and
        Krishna Shankaranarayanan\inst{2} \and
        Ashutosh Trivedi\inst{1} \and
        Namrita Varshney\inst{2}}
\institute{University of Colorado Boulder\email{\{john.komp,ashutosh.trivedi\}@colorado.edu} \and
        Indian Institute of Technology Bombay, Mumbai, India\email{\{abhayrajput,akg,krishnas,ivarnam\}@cse.iitb.ac.in}}

\maketitle
\begin{abstract}
This paper investigates whether recent advances in Large Language Models (LLMs) can assist in translating human explanations into a format that can robustly support learning Linear Temporal Logic (LTL) from demonstrations.
Both LLMs and optimization-based methods can extract LTL specifications from demonstrations; however, they have distinct limitations.
LLMs can quickly generate solutions and incorporate human explanations, but their lack of consistency and reliability hampers their applicability in safety-critical domains. 
On the other hand, optimization-based methods do provide formal guarantees but cannot process natural language explanations and face scalability challenges.
We present a principled approach to combining LLMs and optimization-based methods to faithfully translate human explanations and demonstrations into LTL specifications. 
We have implemented a tool called $\ourtool$ based on our approach. 
Our experiments demonstrate the effectiveness of combining explanations with demonstrations in learning LTL specifications through several case studies.
\end{abstract}


\section{Introduction}
\label{sec:intro}
Linear Temporal Logic~\cite{baier2008principles} (LTL) is a succinct, unambiguous, and expressive formalism that captures the properties of the dynamic evolution of reactive systems. 
LTL specifications are widely supported across various formal methods infrastructure from model checking~\cite{duret2016spot,kvretinsky2018owl} and synthesis~\cite{bohy2012acacia+}, to specification mining~\cite{lemieux2015general,li2011mining} and reinforcement learning~\cite{sadigh2014learning,camacho2019ltl,hahn2019omega}.
However, eliciting an LTL specification largely remains a manual effort, demanding both substantial familiarity with the system and a deep understanding of logic. 
Consequently, capturing complex requirements in LTL remains time consuming and error prone~\cite{whalen2012your}.
For these reasons, automated approaches for extracting LTL specifications from demonstrations are an active area of research~\cite{lemieux2015general,DBLP:conf/atal/AfzalGGS0V23,de2015synthesis}.

The existing approaches for extracting LTL specifications from demonstrations reduce the problem to optimization-based search within a class of LTL formulas using some quantitative semantics of LTL. 
A key challenge in learning LTL from demonstrations is that the demonstrations are finite length traces, while LTL is interpreted over $\omega$-traces.
Consequently, there may be several explanations for a given set of demonstrations. 
For instance, the following demonstrations  $\seq{\set{p},\set{p},\set{p}}$ can be explained as $\globally p$ (globally $p$), $\eventually p$ (eventually $p$), or as $p \wedge \Next p \wedge \Next \Next p$ ($p$ for the first $3$ time steps).
In deciding among competing hypotheses, learning algorithms often use rule-of-thumb regularization such as Occam's razor~\cite{blumer1987occam,domingos1999role,bensusan1998god} in identifying their output. 
We argue that elicitation of LTL formulae from demonstrations can be greatly aided by integrating human gestalt\footnote{The Gestalt principles \cite{gestalt3,gestalt4} refer to insights into how humans recognize patterns, group objects, extend incomplete data, and explain complex concepts based on principles such as {\it continuity}, {\it proximity}, {\it similarity}, and {\it common fate}. 
For instance, a designer, when faced with a sequence $\mathtt{qpqpqpqpppp}$, which might be succinctly explained as a sequence ending with a $\mathtt{p}$, may interpret this sequence by applying the principle of continuity to explain that the system eventually settles in $\mathtt{p}$.} in form of explanations. 
Moreover, the recent success of large language models (LLMs) in understanding and generating natural languages~\cite{openai2023gpt,NIPS2017_3f5ee243,radford2018improving} makes this approach timely and possible.
We study the problem of extracting LTL from demonstrations and explanations to improve the scalability and relevance of the produced LTL formulas.
We leverage LLMs to translate explanations into a form that can be exploited to extract LTL from demonstrations.

\begin{figure}[t]
    \centering
    \begin{center}
    \begin{tikzpicture}[shorten >=1pt,node distance=2.8cm,on grid,thick]
      \node[text width=1cm, align=center] (q0) {};
       \node[text width=1cm,  yshift=3mm, align=center] (q01) {Traces};
      \node[text width=1cm, yshift=-4mm,align=center] (q00) {Expla-\\nation};
 
      \node[draw,text width=1.3cm,align=center,xshift=-7mm,fill=yellow!30] (q1) [right=of q0] {Large language model};
       \node[cloud, yshift=-1.2cm, align=center, cloud puffs=20, cloud ignores aspect, minimum width=1.5cm, minimum height=1cm, draw] (q0p) [above=of q1] {~training prompts~};

      \node[draw,text width=1.5cm,align=center,xshift=6mm,fill=blue!30] (q2) [right of= q1] {Template generation};
      \node[draw,text width=1cm,align=center,xshift=5mm,fill=green!30] (q3) [right of= q2] {Fitness Repair};
      \node[text width=1.5cm,align=center,xshift=-8mm] (q4) [right of= q3] {Synthesized Formula $\phi$};

      \path[->] (q01) edge  node [above] {} ([yshift=8]q1.west);
      \path[->] (q00) edge  node [above] {} ([yshift=-10]q1.west);
      \path[->] (q1) edge  node [above] {$\seq{\varphi_1, {\ldots}, \varphi_k}$} node [below,yshift=-1mm] {$\stackrel{LTL}{Candidates}$} (q2);
      \path[->] (q2) edge  node [above] {$\seq{\tau_1, {\ldots}, \tau_n}$}node [below] {$\stackrel{LTL}{Templates}$} (q3);
      \path[->] (q3) edge  node [above] {} (q4);
      \path[->] (q0p) edge  node [above] {} (q1);

    \end{tikzpicture}  \end{center}\par

    \caption{Our method for synthesizing LTL formulas from traces uses the intuition of an LLM to find possible answers and we correct them using our fitness-aware search.}
    \label{fig:overview}
\end{figure}
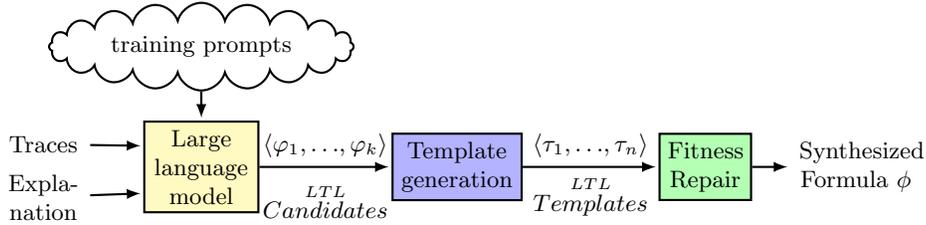

While both LLMs and optimization-based approaches can be used to extract LTL specifications, they have complementary strengths.
While LLMs can integrate both demonstrations and explanations, identify saliency using attention mechanism~\cite{NIPS2017_3f5ee243}, and return unique and novel hypotheses, they often suffer from hallucinations~\cite{zhang2023siren}, returning fictional and untrustworthy answers.  
On the other hand, optimization-based approaches cannot utilize natural language explanations and suffer from scalability challenges.  
To overcome these challenges, we propose integrating both demonstrations and natural language explanations in extracting LTL specifications. 

\vspace{0.4em}\noindent\textbf{Our Approach.} 
We employ quantitative semantics for LTL formulas over finite traces as a quantitative measure of a formula's fitness within the set of demonstrations. 
This fitness measure is used not only to assess the quality of hypotheses generated by an LLM but also to improve the fitness of the resulting formulas through formula repair.
Our high-level approach is outlined in Figure~\ref{fig:overview}. 
We input demonstrations in the form of system traces involving a set of atomic propositions, along with natural language explanations provided by the designer. 
We assume that salient atomic propositions have already been incorporated in the explanations.
After presenting some general prompts regarding LTL syntax, the LLM generates a set of candidate formulas, denoted as $\seq{\phi_1, \ldots, \phi_k}$. 
We evaluate their fitness using our quantitative semantics, and if any of the formulas pass a specified threshold, we accept the corresponding hypothesis. 
However, if none of the generated formulas are deemed suitable, we extract LTL subformulas from this family, which serve as templates $\seq{\tau_1 \ldots, \tau_k}$.
These formulas are subsequently input into our fitness repair algorithm, in conjunction with the demonstrations, to synthesize a formula that offers a more satisfactory explanation of the provided demonstrations.

\begin{figure}[t!]
    \centering
    \includegraphics[width=1\textwidth]{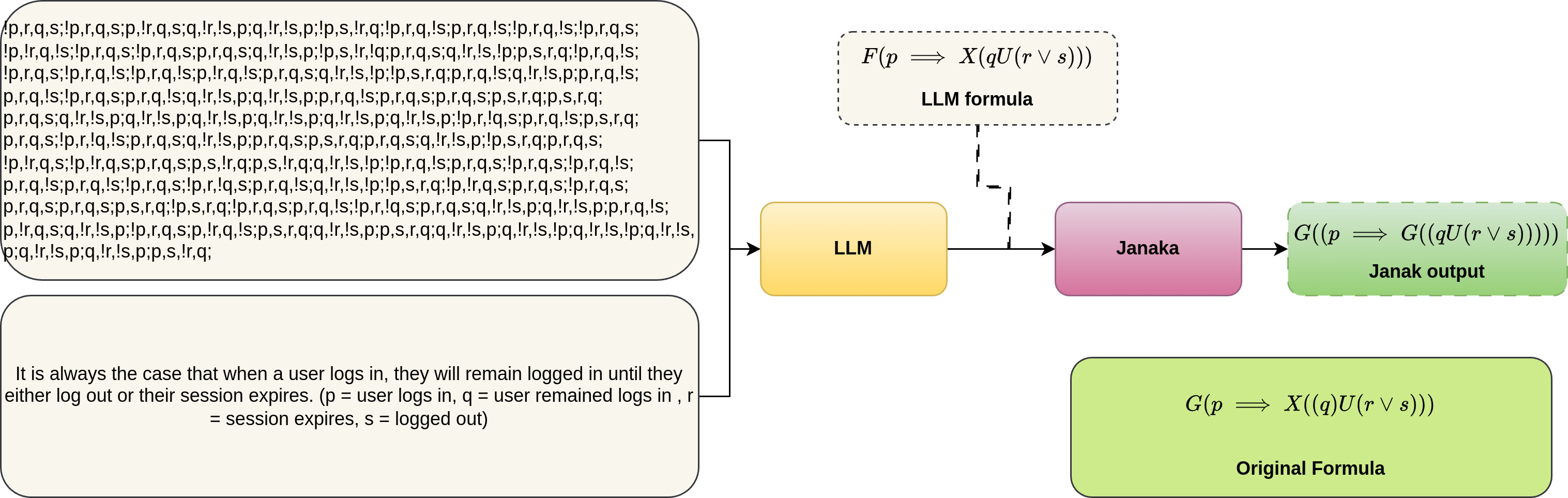}
    \caption{Learning LTL specifications from explanations and demonstrations.}
    \label{fig:mot-example}
\end{figure}

\vspace{-1em}
\subsubsection{Motivating example.} We present a sample use case of our approach through a simple example. Let's consider a scenario where a web designer is keen on formalizing the requirements for an app. 
This formalization should be extracted from the following  set of session logs with semicolon as a separator:
\begin{equation*}
\mathtt{...!p,r,q,s;!p,r,q,s;p,!r,q,s;q,!r,!s,p;q,!r,!s,p;!p,s,!r,q;!p,r,q,!s;...}
\end{equation*}
Here, the symbol $\mathtt{p}$ stands for a user logging in, $\mathtt{q}$ for a user continuing to use the app, $\mathtt{r}$ for the expiration of the session, and $\mathtt{s}$ for a user logging out. While from this limited information, it is difficult to formalize the behavior, the designer can provide the following description:
\begin{quotation}
\noindent\emph{It is always the case that when a user logs in, they will remain logged in until they either log out or their session expires.}
\end{quotation}
$\chatgpt$, with some minimal LTL syntax prompts, demonstrations, and explanations, can produce the following explanation $\finally (p \implies \nextt(q \until (r \lor s)))$.

It is clear to see that the description does not match the behaviour expressed in the explanations. Moreover, the session logs do not match this specification as well.
On the other hand, it has correctly predicted the saliency of $(r \lor s)$, $q \until (r \lor s)$ and an implication of predicate $p$.
$\ourtool{}$ is able to create templates from this formula and invoke constraint solvers to discover an LTL formula providing an improved fitness.
The following LTL formula was generated in the process
$\globally (p \implies \globally(q \until (r \lor s)))$,
that arguably and quantitatively provides a better fit both for the natural language explanation and the set of demonstrations.
This interaction, along with the the ground truth LTL formula, is shown in Figure~\ref{fig:mot-example}.

This approach has been implemented in a Python-based tool, $\ourtool$, to synthesize LTL formulae from demonstrations and explanations. The tool utilizes $\gurobi$~\cite{gurobi} for constraint solving and can be interfaced with $\chatgpt$, $\claude$, or $\bard$. 
However, for our experiments we report results using $\chatgpt$ by OpenAI \cite{openai2023gpt} as an LLM model.
Unless explicitly stated, we use the term LLM to refer to $\chatgpt$. 
We conducted experiments over 17 case studies, providing both demonstrations and natural language explanations of the underlying system. The goal of our experiments is to address the following research questions:
1) Can LLMs be used for learning LTL from demonstrations? 2) Can human explanations increase the reliability of hypotheses generated by LLMs? and 3) Can we extract useful LTL patterns from the outputs of LLMs to construct LTL formulas more efficiently than learning LTL directly from demonstrations?

Our experiments demonstrate the effectiveness of combining explanations with demonstrations in learning LTL specifications.
We applied the LLM and tool combination on a representative set of formulas and experimental results suggest that our approach produces better quality formulas for 12 cases than solely using an LLM. 
We also show that using traces improves the responses of LLM for 10 cases and in three cases the response remained the same.

\section{Problem Definition} 
\label{sec:prelims}
For a predicate $p$ we write $[p] = 1$ if $p$ is true and $0$ otherwise. 
An \emph{alphabet} $\Sigma$ is a non-empty, finite set of \emph{symbols}. 
A \emph{finite word} $w$ over $\Sigma$ is a finite
sequence $\seq{a_1 a_2 ... a_n}$ of symbols from $\Sigma$. 
The empty sequence is called the empty word, denoted $\epsilon$. 
An $\omega$ word $w$ over $\Sigma$ is a function $w: \mathbb{N} \to \Sigma$.
The domain of $w$, denoted $\mathsf{dom}(w)$ is the set of positions in $w$. 
% Thus, $\mathsf{dom}(a_1 \dots a_n)=\{1,2,\dots,n\}$ and $\mathsf{dom}(\epsilon)=\emptyset$. 
The length of a finite (or infinite) word $w$ is denoted $|w|$ and is equal to the size of its domain.  
For a  word $w=a_0a_1\ldots$, and 
two indices $i \leq j \leq |w|$, we write $w[i,j]$ to denote the sub-word $a_ia_{i+1} \dots a_j$.
and $w[i]$ for the suffix of $w$ starting at index $i$.
Abusing notation, we write a finite word $w = a_0a_1\cdots a_n$ over $\Sigma$ as a partial function $w: \mathbb{N} \to \Sigma$ where $w(i) = a_i$ for $i \leq n$ and $w(i)$ is undefined for $i > n$. 
Let $\Sigma^*$ and $\Sigma^\omega$ be the set finite and infinite words over $\Sigma$, respectively.

\subsection{Qualitative Semantics for LTL}
Let $\prop$ be 
a set of propositional variables. 
LTL~\cite{baier2008principles} is an
extension of propositional logic with temporal modalities.

Formulas in LTL over $\prop$ are defined by the following grammar:
\begin{equation}
\nonumber
\varphi ::= \true ~|~a \in \prop~|~\varphi_1 \vee \varphi_2~|~\neg
\varphi~|~\Next~\varphi~|~\varphi \until \varphi
\end{equation}
Using the above basic operators, other Boolean connectives such as $\varphi \wedge \psi = \neg (\neg \varphi \vee \neg \psi)$ and $\varphi {\implies} \psi = \neg \varphi \vee \psi$, as well as temporal operators (\emph{finally}) $\finally \varphi= \true \until
\varphi$ and (\emph{globally}) $\globally \varphi=\neg \finally \:\neg \varphi$ can be derived as syntactic sugar.

Let $\Sigma=2^{\prop}$. The semantics of LTL over the set of propositions $\prop$ is defined over words in $\Sigma^\omega$. 
The notation $w \models \varphi$ denotes that the word $w$ satisfies the formula 
$\varphi$, while $w \not \models \varphi$ denotes that it does not.
\begin{itemize}
    \item $w[i] \models \true$.
    \item $w[i] \models a$ if and only if $a \in w(i)$. 
    \item $w[i] \models \neg \varphi$ if and only if $w[i] \not \models \varphi$.
    \item $w[i] \models \varphi \vee \psi$ if and only if
    $w[i] \models \varphi$ or $w[i] \models \psi$.
    \item $w[i] \models \Next \varphi$ if and only if $i+1 \leq |w|$ and $w[i+1] \models \varphi$. 
    \item $w[i] \models \varphi \until \psi$ if and only if there exists $j \geq i \in \dom{(w)}$ such that $w[j] \models \psi$ and for all $i \leq k < j$, $w[k] \models \varphi$.
\end{itemize}
We say $w \models \varphi$ if $w[0] \models \varphi$.
The language of an LTL formula $\varphi$ is defined as
$L(\varphi) = \{w \in \Sigma^\omega \mid  w \models \varphi\}$.
We say that two LTL formulas are language equivalent if they accept the same language. 

LTL formulas $\finally \globally p$  (eventually forever $p$) and $\globally \finally p$  (infinitely often $p$) are not $\omega$-language equivalent, but they are finite-language equivalent~\cite{de2013linear} capturing the set of finite words ending with a $p$.

\subsection{Quantitative semantics for LTL}
\label{sec:robust}

Given an LTL formula $\varphi$ and a finite trace, we design two valuation functions $\sem{\varphi, w}$ that quantify the fitness of $\varphi$ in explaining $w$. These valuation functions are based on two recently introduced quantitative semantics: \semanticsone \cite{DBLP:conf/atal/AfzalGGS0V23} and \semanticstwo \cite{alur2023policy}. 
We support both semantics to cater to two different intuitions behind the synthesis problem that needs to be solved.
The robust semantics returns a positive value when a trace satisfies a formula and a negative value otherwise. Moreover, higher positive and negative values convey the degree to which a perturbed trace satisfies the formula. 
The discounted semantics returns a value in the range $[0, 1]$, characterizing the discounted satisfaction.

The complexity of an LTL formula can result from two aspects: the complexity of the temporal structure ($\globally p$ is simpler than $p \wedge \Next p \wedge \Next \Next p$ in explaining the sample $\seq{\set{p}\set{p}{\set{p}}}$ and the complexity of the nesting structure (the formula $p$ is simpler than $p \wedge \neg q$ for explaining the sample $\seq{\set{p}}$. 
We use the temporal discounting parameter  $\alpha$ and nesting discount factor $\beta$ in weighing this source of complexity.  
We modify the semantics of \cite{DBLP:conf/atal/AfzalGGS0V23} and ~\cite{alur2023policy} to integrate these parameters. 
\vspace{-1em}
\subsubsection{\semanticsone.}
Let $0 {\leq} \alpha {\leq} 1$ and $0 {\leq} \beta {\leq} 1$ be temporal and nesting discount factors.
In the robust semantics \cite{DBLP:conf/atal/AfzalGGS0V23}, given an LTL formula $\varphi$ and a finite word $w \in \Sigma^*$, a 
 valuation function  $\sem{\varphi, w}$ is defined as follows. 
\begin{itemize}
\small
    \item
    $ \sem{a,w} = 2\cdot[a \in w(0)]-1$ and 
    $\sem{\neg a, w} = -1\cdot \sem{a,w}$
\small
    \item 
    $
    \sem{\varphi \land \psi, w} =
    \begin{cases}
        \beta \cdot \sem{\varphi, w} \cdot \sem{\psi, w}   & \text{if~}  \sem{\varphi, w} \geq 0 \land \sem{\psi, w} \geq 0, \\
        -1 & otherwise
    \end{cases}
    $
    \item
    $
    \sem{\varphi \lor \psi, w} =  \beta \cdot
    \begin{cases}
     \mathrm{avg}(\sem{\varphi, w},\sem{\psi, w}) &      
     % \frac{(\sem{(\varphi, w)}+ \sem{(\psi, w)})}{2} & 
     \text{if~}  \sem{\varphi, w} \geq 0 \wedge  \sem{\psi, w} \geq 0,\\
 \max(\sem{\varphi, w}, \sem{\psi, w}) &  \text{otherwise}   
    \end{cases}
    $
\small
    \item %Globally ($G$): 
    $
    \sem{G\varphi, w} = \beta \cdot
    \begin{cases} 
     \displaystyle\sum_{i=0}^{|w|} \alpha^i \sem{\varphi, w[i]} & \text{if } \sem{\neg\varphi, w[t]}< 0, \text{ for all $t \leq |w|$} , \\
    -1 & \text{otherwise}.
\end{cases}
    $
    \item %Finally ($F$): 
    $
    \sem{F\varphi, w} = \beta \cdot
    \begin{cases} 
     \alpha^t \sem{\varphi, w[t]} & \text{where } t = \min\{j | \sem{\varphi, w[j]} \geq 0\}, \\
    \gamma\cdot\alpha^{|w|} & \text{if } \sem{\varphi, w[t]} = 0 ~\text{for all}~ t.
\end{cases}
    $
    \item 
    $
    \sem{(X\varphi, w)} = 
    \begin{cases} 
          \sem{\varphi, w[1]} & \text{if } \sem{\varphi, w[1]}\geq0  
          \land |w|>1  \\
          \gamma & \text{if~} |w|<2 \\
        -1 & \text{otherwise}.
    \end{cases}
    $
    \item
    $
    \sem{\varphi U \psi, w} = 
    \begin{cases} 
       \alpha^t \cdot \sem{\psi, w[j]} & t=\min\{j | \sem{\psi, w[j]} \geq 0\}
      \text{ and } \sem{\varphi, w[i]} \geq 0  \text{ for all } i < t,~\\
      \alpha^{|w|} \cdot \gamma  & \text{if }  \sem{\psi, w[i]} \geq 0 \text{ for all $i \leq |w|$}\\
     -1 & \text{ otherwise}.
\end{cases}
    $
\end{itemize}
\begin{proposition}[Qualitative Equivalence: Robust]
If $\alpha = 1$ then for every finite trace $w \in \Sigma^*$, we have that $\sem{\varphi, w} = 1$ if and only if $w \models \varphi$.
\end{proposition}
\subsubsection{\semanticstwo.}
Let $0 {\leq} \alpha {\leq} 1$ and $0 {\leq} \beta {\leq} 1$ be temporal and nesting discount factors.
For a finite trace $w$ and LTL formula $\varphi$, we define the valuation of $\varphi$ wrt $w$,  $\sem{\varphi, w}$ as follows:
\begin{itemize}
    \item $\sem{\true, w} = 1$.
    \item $\sem{a, w} = 1$ if and only if $w(0) \ni a$.
    \item $\sem{\neg \varphi, w} = 1 - \sem{\varphi, w}$.
    \item $\sem{\varphi \vee \psi, w} = \beta \cdot \max \set{\sem{\varphi, w}, \sem{\psi, w}}$.
    \item $\sem{\Next \varphi, w} = \alpha \sem{\varphi, w[1]}$.
    \item $\sem{\varphi \until \psi, w} = \max\limits_{0 \leq i < |w|} \set{\min \set{\alpha^i \cdot \sem{\psi, w[i]}, \min\limits_{0 \leq j < i} \set{\alpha^j \cdot \sem{\varphi, w[i]}}}}.$
\end{itemize}
For $\globally$ and $\eventually$ operators, these semantics follow from those of $\until$ and given here for completeness: 
\begin{itemize}
    \item $\sem{\eventually \varphi, w} = \max\limits_{0 \leq i < |w|} \set{\alpha^i \cdot \sem{\varphi, w[i]}}.$
   \item 
   $\sem{\globally \varphi, w} = 1 - \max\limits_{0 \leq i < |w|} \set{\alpha^i \cdot \sem{\neg \varphi, w[i]}} = 1 - \max\limits_{0 \leq i < |w|} \set{\alpha^i \cdot ( 1 - \sem{\varphi, w[i]})} .$
\end{itemize}

These quantitative semantics are similar to discounted semantics defined in~\cite{alur2023policy} except that our semantics introduce an additional discounting for Boolean operators and they have been defined for finite traces.
\begin{proposition}[Qualitative Equivalence: Discounted]
    If $\alpha = 1 $ and $\beta = 1$ then for every finite trace $w \in \Sigma^*$, we have that $\sem{\varphi, w} = 1$ if and only if $w \models \varphi$.
\end{proposition}

\subsection{Mining LTL from Demonstrations and Explanations}
A \emph{sample}, or set of demonstrations, $\sample$ is a   finite set of traces $P \subseteq (2^{\prop})^*$. 
In this work, depending on the context, we use the term trace and demonstration interchangeably. 
For an LTL formula \(\varphi\), we say \(\sample 
\models \varphi\) iff for every $\tau \in P$ we have that $\tau \models \varphi$. 
Given a sample $\sample$, we are interested in learning an LTL formula $\varphi$ such that $\sample \models \varphi$. 
While the qualitative criterion for fitness may not be selective enough, as there may be multiple LTL hypotheses that satisfy a given sample, we argue that instead of making decisions based on Occam's razor principles, it is arguably better to integrate designer's insight in selecting competing hypotheses. For this reason, we study the following problem.

\begin{definition}[Qualitative Mining for LTL]
Given a set of demonstrations $\sample$, and a natural language explanation $\mathcal{E}$ of the system capturing designer's insights, synthesize an LTL formula $\varphi$ such that $\sample \models \varphi$. 
\end{definition}

Constraint solvers such as $\gurobi$ can be used to synthesize LTL formulas from demonstrations; however, they are unable to process natural language explanations. LLMs, on the other hand, can work with both explanations and demonstrations, but do not provide formal guarantees. We posit that LLMs can often generate a set of competing hypotheses such that while none of the proposed formulas may capture the specification in its entirety, together they capture different salient features from the explanation. To utilize this feature of LLMs, while providing formal guarantees, we use constraint solvers to mix-and-match features from this set of formulas to generate the final formula.

To measure the partial fitness of a formula over demonstrations and to aid in combining the best-performing formulas, we use quantitative semantics defined in the previous section as a fitness measure. For a given semantics $\sem{\cdot}$, a sample $\sample$, the valuation of $\varphi$ with respect to $\sample$ is defined as the average of all valuations over all traces in the sample, i.e., $\sem{\varphi, \sample} = (1/|w|)\cdot \sum_{w \in \sample}^{} \sem{\varphi, w}$. We will refer to $\sem{\varphi, \sample}$ and $\sem{\varphi, w}$ as the \emph{fitness} of $\varphi$ with respect to the sample $\sample$ and $w$, respectively. Our approach tackles a quantitative relaxation of the previous problem.

\begin{definition}[Quantitative Mining for LTL]
Given a set of demonstrations $\sample$, and a natural language explanation $\mathcal{E}$ of the system capturing designer's insights, and a bound $\kappa$, learn an LTL formula $\varphi$ such that $\sem{\varphi, \sample} \geq \kappa$. 
\end{definition}

\section{Quantitative Mining for LTL}
\begin{figure}[t!]
    \centering
    \begin{minipage}{.025\textwidth}
        \centering
        \begin{tikzpicture}[-,>=stealth',shorten >=1pt,auto,node distance=1cm,semithick]
          \node[] (A)    at (0,2)             {$\globally$};
          \node[] (B)    at (0, 1.4)           {$p$}; 
         \path (A)  edge node {} (B) ;
        \end{tikzpicture}
        \label{fig:Temp1}
    \end{minipage}
    \hspace{0.005\textwidth}
    \begin{tikzpicture}
              \node[] (0,0) {};
              \draw[->,thick] (0,0.25) -- (0.5,0.25) node [above=1cm, left] {(1)};
    \end{tikzpicture}
    \begin{minipage}{.05\textwidth}
        \centering
        \begin{tikzpicture}[-,>=stealth',shorten >=1pt,auto,node distance=1cm,semithick]
          \node[] (A)    at (0,2)             {$\phi$};
          \node[] (B)    at (-0.5, 1.4)           {$p$}; 
          \node[] (C)    at (0.5, 1.4)           {$\phi$}; 
         \path (A)  edge node {} (B) ;
          \path (A)  edge node {} (C) ; 
        \end{tikzpicture}
        \label{fig:Temp2}
    \end{minipage}
    \hspace{0.05\textwidth}
    \rule{0.5pt}{1.cm}
    \begin{minipage}{.025\textwidth}
        \centering
        \begin{tikzpicture}[-,>=stealth',shorten >=1pt,auto,node distance=1cm,semithick]
          \node[] (A)    at (0,2)             {$\globally$};
          \node[] (B)    at (0, 1.4)           {$p$}; 
         \path (A)  edge node {} (B) ;
        \end{tikzpicture}
        \label{fig:Temp3}
    \end{minipage}
    \begin{tikzpicture}
              \node[] (0,0) {};
              \draw[->,thick] (0,0.25) -- (0.5,0.25) node [above=1cm, left] {(2)};
    \end{tikzpicture}
    \begin{minipage}{.025\textwidth}
        \centering
        \begin{tikzpicture}[-,>=stealth',shorten >=1pt,auto,node distance=1cm,semithick]
          \node[] (A)    at (0,2)             {$\globally$};
          \node[] (B)    at (0, 1.4)           {$\phi$}; 
         \path (A)  edge node {} (B) ;
        \end{tikzpicture}
        \label{fig:Temp4} 
    \end{minipage}
    \rule{0.5pt}{1cm}
    \begin{minipage}{.05\textwidth}
        \centering
        \begin{tikzpicture}[-,>=stealth',shorten >=1pt,auto,node distance=1cm,semithick]
          \node[] (A)    at (0,2)             {$\vee$};
          \node[] (B)    at (-0.5, 1.4)           {$p$}; 
          \node[] (C)    at (0.5, 1.4)           {$q$}; 
         \path (A)  edge node {} (B) ;
          \path (A)  edge node {} (C) ; 
        \end{tikzpicture}
        \label{fig:Temp5}
    \end{minipage}
    \hspace{0.05\textwidth}
    \begin{tikzpicture}
                \centering
              \node[] (0,0) {};
              \draw[->,thick] (0,0.25) -- (0.5,0.25) node [above=1cm, left] {(3)};
    \end{tikzpicture}
    \hspace{0.005\textwidth}
    \begin{minipage}{.05\textwidth}
        \centering
        \begin{tikzpicture}[-,>=stealth',shorten >=1pt,auto,node distance=1cm,semithick]
          \node[] (A)    at (0,2)             {$\phi$};
          \node[] (B)    at (-0.5, 1.4)           {$p$}; 
          \node[] (C)    at (0.5, 1.4)           {$q$}; 
         \path (A)  edge node {} (B) ;
          \path (A)  edge node {} (C) ; 
        \end{tikzpicture}
        \label{fig:Temp6}
    \end{minipage}
\caption{Template Generation Methods}
\label{fig:TemplateGen}
\end{figure}
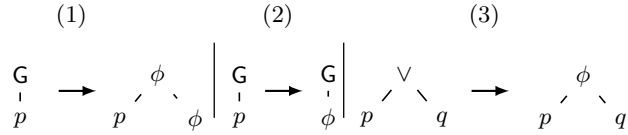

 Our approach to generate and repair LTL from LLMs is shown in Algorithm~\ref{algo:llm_ltl}. 
 The input to the LLM consists  of a set $T$ of traces, a natural language description of the formulas, and a fixed prompting text for LTL syntax.
We prompt the LLM to output a fixed number of formulae $\psi_1, \dots, \psi_N$ (Algorithm \ref{algo:llm_ltl}, line \ref{formulaegen}), evaluate them based on the valuation function described above, and if satisfactory (fitness beyond the tolerance) return the best fit formula. Otherwise, we compute the top $k$ fit formulas (line \ref{chooseform}) and generate templates from these formulas using Algorithm~\ref{algo:template_gen_clear} as follows.
From the parse tree of $\Psi_{top}$ ($Tree(\Psi_{top}$))  we synthesize a \emph{template} as shown in Figure \ref{fig:TemplateGen}. A template is obtained by randomly choosing a node, say $v$, 
in $Tree(\Psi_{top})$ and replacing it with a tree
of height $\leq d$ for some chosen $d \geq 1$ while preserving the following: 
\begin{enumerate}
    \item If 
$v$ has only one child in $Tree(\Psi_{top})$, then 
the template hides the label of node $v$ making it a hole which can be filled,  and also adds a tree $t'$ as its right child. 
All nodes of $t'$ are holes, which need to be filled, and its height is $\leq d$.
\item If $v$ has two children in $Tree(\Psi_{top})$, then the template only hides the label of $v$ making it a hole to be filled. 
\item If $v$ is a leaf node in $Tree(\Psi_{top})$, then 
the template makes the label of $v$ a hole and adds a tree $t'$ of height $\leq d$ at $v$, with all nodes of the newly added tree $t'$ being holes to be filled. 
\end{enumerate}
 Thus, we replace $v$ with a \emph{template hole} whose depth is $\leq d$. 
 Note that the choice of the node $v$ is non-deterministic (line \ref{random} in Algorithm \ref{algo:template_gen_clear}). 
  Thus, Algorithm \ref{algo:template_gen_clear} 
starts with $\Psi_{top}$, and a chosen $d \geq 1$, and outputs a new parse tree containing template holes. The height of the new tree is $\leq d+ \textsf{height}(Tree(\Psi_{top})$.  
In line \ref{templategen} of Algorithm \ref{algo:llm_ltl}, 
let $\Psi_{temp}$ be the parse tree generated using Algorithm \ref{algo:template_gen_clear}.

 \begin{algorithm}[t!]
    \caption{$\ourtool$(Traces T, Explanations E, Prompt P, d, $\kappa$)}
    \label{algo:llm_ltl}

    $\Psi \leftarrow$ \textsc{API2LLM}(T, E, P)\label{formulaegen}    \tcp*{API call to LLMs}
    
    \lIf{$\max_{\psi \in \Psi} \{ \sem{\psi,T} \geq \kappa$}{\Return $\argmax_{\psi \in \Psi} \{ \sem{\psi,T}\}$}
    \Else{
      $\Psi_{top} \leftarrow \displaystyle\mathrm{top\_k}_{\psi \in \Psi} \{ \sem{\psi,T}$ \}   \label{chooseform}\tcp*{Returns top $k$ fit formulas}
    
    $\Psi_{temp} \leftarrow$ \textsc{TemplateGen}($\Psi_{top}$,d)\label{templategen}    \tcp*{Generate templates}
      
    \Return $\textsc{LTLRepair}(T, \Psi_{temp} , d, \kappa)$    \tcp*{Repair LTL from templates}}  
\end{algorithm}

\begin{algorithm}[t]
    \caption{\textsc{TemplateGen}(Set of LTL Formulas $\Psi$, Depth d)}
    \label{algo:template_gen_clear}
    
    Combine the given formulas $\Psi$ to construct a parse tree Tree($\Psi$)\;
    
    \For{ each visit to a node of Tree($\Psi$) in breadth-first order}{
        \If{$\textsc{random}(0,1)< 0.2$}{ \label{random}
            Replace the node in Tree($\Psi$) by a template hole of depth $\leq d$
        }
    }
    \Return{Tree($\Psi$)}\;
    
\end{algorithm}

\begin{algorithm}[t]
    \caption{\textsc{LTLRepair}( Traces $T$, Templates $\Psi$, Depth $d$, $\kappa$)}
    \label{algo:karak3}
        Construct tree constraint $\Phi^{template}_d$ from $\Psi$ \tcp*[r]{Constraints in eq.(\ref{eq:lt1})-(\ref{eq:lt10})}
    Construct trace constraint $\Phi^T_d$ \tcp*[r]{Constraints in eq.(\ref{eq:label_score_Disc})-(\ref{eq:Untill_Disc})}
    $\mathtt{fitness},m \leftarrow \max \left( \displaystyle\sum_{\tau=1}^{|X|} y_{1,0}^\tau \: |\: \tau \in T \right) $ such that $m \models \Phi^{template}_d \land \Phi^T_d$
    
    \lIf{$\mathtt{fitness} \geq \kappa$}{
        \Return the formula constructed from m
    }
    \lElse{
        \Return False
    }
    
\end{algorithm}

\subsection{LTL Repair Algorithm}
Next, we use Algorithm \ref{algo:karak3} which takes as input the modified parse tree with template holes, $\Phi_{temp}$, to synthesize 
an optimal formula with the highest score given $\Phi_{temp}$, the set of traces $T$ and 
the depth $d$ which was used in the generation of the template holes.  The optimal formula is obtained by computing and optimizing
the scores of a class of formulas whose  depth is at most $d+ \textrm{depth}(\Phi_{top})$, and is  constrained to be well formed
and satisfied by traces $T$. This is done by building a constraint system and finding an optimal solution for it. The constraint system $\Phi= \Phi^{template}_d \wedge \bigwedge_{\tau \in T}\Phi^{\tau}_d$ is obtained as a conjunction of the 
constraints obtained by encoding the template ($\Phi^{template}_d$) as well as the traces ($\bigwedge_{\tau \in T}\Phi^{\tau}_d$). 
The first part encodes the structure of the \emph{template}, while the second part encodes the
\emph{functional} constraints based on the on the \emph{score} of each node enforced by the operators in the formula.  The template constraints are uniform across both the robust and discounted semantics, while the functional 
constraints depend on the chosen semantics. 

 To encode the formula structure, we use a syntax tree with  identifiers
$\nodes=\{1, 2, \ldots, n\}$ for the set of nodes, where $n = |\nodes|$.
We assume herein that the root node is
identified as $1$.  We have the child relations $L$ and $R$, such that $(i, j)
\in L$ (or $R$) iff the node $j$ is the left (right) child of node $i$. The only child of unary operators  
is considered as a  left child by assumption. For each node $i \in \nodes$ and
possible label $\lambda \in \op \;\union\; \prop$, we introduce a Boolean variable
$x_{i, \lambda}$ indicating whether the node is labelled with an operator
($\op$) or variable ($\prop$).
We consider $x_{i, l}^\tau$ as a Binary variable where $i \in \text{nodes}$ and $l \in \prop \cup \op$ and $\tau$ is a trace. $\op$ and $\prop$ are the sets of all operations and propositions, respectively. Let  $Br$ be a binary operation ($\land$, $\lor$, $U$, $\implies$), $Ur$  denote a unary operator ($G$, $F$,$X$), and $Lf$ denote a  leaf node labeled by a proposition $\prop$).

\noindent{\bf{Depth One tree Constraints}}. In the case of a tree whose depth is one, we have only a leaf node labeled by the propositional variables. Exactly one propositional variable must be decorating a leaf node; this is captured by 
{\footnotesize
\begin{gather}
   \sum _{\substack{op \in Lf}} x^\tau _{i, op} = 1   \label{eq:1+LF}
\end{gather}
}%
\noindent{\bf{Tree Constraint(depth ``d'' $>$ 1)}.} In the case of trees with a depth of more than one, the constraints are formed based on whether each non-leaf node is decorated by a unary or a binary operator. The constraints are a case analysis depending on the operator labeling the nodes. 

\noindent For every non-leaf node $j$ of parent node $i$, i.e., $(i,j) \in L$ or $(i, j) \in R$, we have the following constraints:
{\footnotesize
\begin{gather}
  \bigwedge\limits _{\substack{1 \leq i \leq 2^{d-1} - 1\\(i, j) \in L \union R }}\; \sum _{\substack{op_i \in \text{Br}  }} x^\tau _{i, op_i} = 1 \implies \sum _{\substack{op_j \in \op}} x^\tau _{j, op_j} = 1    \label{eq:lt1} \\
  \bigwedge\limits _{\substack{1 \leq i  \leq 2^{d-1} - 1\\(i, j) \in L  }}\; \sum _{\substack{op_i \in \text{Ur}  }} x^\tau _{i, op_i} = 1 \implies \sum _{\substack{op_j \in \op}} x^\tau _{j, op_j} = 1   \label{eq:lt2} \\
   \bigwedge\limits _{\substack{1 \leq i \leq 2^{d-1} - 1 \\(i, j) \in R}}\; \sum _{\substack{op_i \in \text{Ur}  }} x^\tau _{i, op_i} = 1 \implies \sum _{\substack{op_j \in \op}} x^\tau _{j, op_j} = 0  \label{eq:lt3} \\
    \bigwedge\limits _{\substack{1 \leq i \leq 2^{d-1} - 1\\(i, j) \in L \union R}}\; \sum _{\substack{op_i \in \text{Lf}  }} x^\tau _{i, op_i} = 1 \implies \sum _{\substack{op_j \in \op}} x^\tau _{j, op_j} = 0   \label{eq:lt4} \\
    \bigwedge\limits _{\substack{1 \leq i \leq 2^{d-1} - 1 \\(i, j) \in L \union R}}\; \sum _{\substack{op_i \in \op  }} x^\tau _{i, op_i} = 0 \implies \sum _{\substack{op_j \in \op}} x^\tau _{j, op_j} = 0   \label{eq:lt5} 
\end{gather}
}%
For every leaf node $j$ of parent node $i$, i.e., $(i,j) \in L$ or $(i, j) \in R$, we have the following constraints:
{\footnotesize 
\begin{gather}
    \bigwedge\limits _{\substack{2^{d-1} \leq j \leq 2^{d} - 1\\(i, j) \in L \union R }}\; \sum _{\substack{op_i \in \text{Br}  }} x^\tau _{i, op_i} = 1 \implies \sum _{\substack{op_j \in Lf}} x^\tau _{i, op_j} = 1\label{eq:lt6} \\
    \bigwedge\limits _{\substack{2^{d-1} \leq j \leq 2^{d} - 1\\(i, j) \in L  }}\; \sum _{\substack{op_i \in \text{Ur}  }} x^\tau _{i, op_i} = 1 \implies \sum _{\substack{op_j \in Lf}} x^\tau _{j, op_j} = 1   \label{eq:lt7} \\
    \bigwedge\limits _{\substack{2^{d-1} \leq j \leq 2^{d} - 1 \\(i, j) \in R}}\; \sum _{\substack{op_i \in \text{Ur}  }} x^\tau _{i, op_i} = 1 \implies \sum _{\substack{op_j \in Lf }} x^\tau _{j, op_j} = 0  \label{eq:lt8} \\
    \bigwedge\limits _{\substack{2^{d-1} \leq j \leq 2^{d} - 1 \\(i, j) \in L \union R}}\; \sum _{\substack{op_i \in \text{Lf}  }} x^\tau _{i, op_i} = 1 \implies \sum _{\substack{op_j \in Lf}} x^\tau _{j, op_j} = 0   \label{eq:lt9} \\
    \bigwedge\limits _{\substack{2^{d-1} \leq j \leq 2^{d} - 1 \\(i, j) \in L \union R}}\; \sum _{\substack{op_i \in \op  }} x^\tau _{i, op_i} = 0 \implies \sum _{\substack{op_j \in \text{Lf}}} x^\tau _{j, op_j} = 0   \label{eq:lt10} 
\end{gather}
}

\subsubsection{Trace Constraints.} Next, we encode the functional constraints imposed by the operators for discounted semantics. 
Trace constraints for the robust semantics are analogous and given in Appendix~\ref{app:rob-cons}. 
For this, to each node $i \in \nodes$, we attach a set of real variables $Y^\tau _i = \{y^\tau _{i, t} \; \mid 0 \leq t \leq |\tau|, i \in \nodes\}$ representing its
 \emph{score} at each point $0 \leq t \leq |\tau|$ in a trace $\tau \in  T$, where $y^\tau _{i, t}$ is defined below.  
  For a given trace $\tau$ we construct $\varphi^\tau _\depth$ as the conjunction of:
{\footnotesize 
\begin{gather}
  0<y^\tau_{1,0} \leq 1  \\
  %\textstyle
    \bigwedge\limits_{1 \leq i \leq n} \bigwedge\limits_{p \in \prop }\;\;x_{i, p} \Rightarrow \left[ 
        \bigwedge\limits_{1 \leq t \leq |T|} y^\tau_{i, t} = \begin{cases}
             1 & ~if~ p \in \tau (t) \\
            0 & ~if~ p \not \in \tau (t) 
        \end{cases}
      \right] \label{eq:label_score_Disc} %\\
\end{gather}
\begin{gather}
      \bigwedge\limits_{1 \leq i \leq n} \bigwedge\limits_{p \in \prop }\;\;x_{i, !p} \Rightarrow \left[ 
        \bigwedge\limits_{1 \leq t \leq |T|} y^\tau_{i, t} = \begin{cases}
             0 & ~if~ p \in \tau (t) \\
            1 & ~if~ p \not \in \tau (t) 
        \end{cases}
      \right] \label{eq:not_l_score_Disc} \\
    \bigwedge\limits_{\substack{1 \leq i \leq n, \\(i, j) \in L, \\(i, j') \in R}}\;\;x_{i, \land} \Rightarrow \left[         \bigwedge\limits_{1 \leq t \leq |T|} y^\tau_{i, t} = \beta\cdot \min\{ y^\tau_{j,t}, y^\tau_{j', t}\}       \right] \label{eq:and_score_Disc} \\
    \bigwedge\limits_{\substack{1 \leq i \leq n, \\(i, j) \in L, \\(i, j') \in R}}\;\;x_{i, \lor} \Rightarrow \left[        \bigwedge\limits_{1 \leq t \leq |T|}  y^\tau_{i, t} = \beta\cdot \max\{ y^\tau_{j,t}, y^\tau_{j', t}\}       \right] \label{eq:or_score_Disc}  \\
       \bigwedge\limits_{\substack{1 \leq i \leq n, \\(i, j) \in L, \\(i, j') \in R}}\;\;x_{i, \implies} \Rightarrow \left[         \bigwedge\limits_{1 \leq t \leq |T|} y^\tau_{i,t} = \beta\cdot \max\{1-y^\tau_{j,t}, y^\tau_{j', t}\} 
      \right] \label{eq:implies_score_Disc} \\    
    \bigwedge\limits_{\substack{1 \leq i \leq n, \\(i, j) \in L}}\;\;x_{i, G} \Rightarrow \left[ 
       \bigwedge\limits_{1 \leq t \leq |T|} y^\tau_{i,t} = \beta \cdot \{ 1- \max\limits_{t\leq t' \leq|T|}           
   \{\alpha^{t'-t} \cdot 1-y^\tau_{j,t'}\}\} 
      \right] \label{eq:Globally_Disc} \\
    \bigwedge\limits_{\substack{1 \leq i \leq n, \\(i, j) \in L}}\;\;x_{i, F} \Rightarrow \left[ 
         \bigwedge\limits_{1 \leq t \leq |T|} y^\tau_{i,t} = \beta \cdot \max\limits_{t\leq t' \leq |T|}
         \{\alpha^{t'-t} \cdot y^\tau_{j,t'}\} 
      \right] \label{eq:Future_Disc} \\
    \bigwedge\limits_{\substack{1 \leq i \leq n, \\(i,j) \in L}}\;\;x_{i, X} \Rightarrow \left[       \bigwedge\limits_{1 \leq t \leq |T|}  y^\tau_{i,t} =   y^\tau_{j,t+1}
      \right] \label{eq:next_Disc} \\
    \bigwedge\limits_{\substack{1 \leq i \leq n}}\;\;x_{i, U} \Rightarrow \left[ 
        \bigwedge\limits_{1 \leq t \leq |T|}   y^\tau_{i,t} = \min \set{\alpha^i \cdot \sem{\psi, w[i]}, \min\limits_{0 \leq j < i} \set{\alpha^j \cdot \sem{\varphi, w[i]}}}
      \right] \label{eq:Untill_Disc} 
\end{gather}
}%

For each Boolean variables and operator, the constraints encode the valuation of a node as a function of the valuation of its children.

The score for a node labelled $\globally$ at a position $t$ in the trace is
 described by the constraint (\ref{eq:Globally_Disc}). We encode the actual score
as a function of its child, adding the child's score
 over all future positions in the input word, scaled by an exponential. 
 We also check that the $\globally$-property holds in a classical
 sense, i.e., that its child has non-negative  valuation at all positions. Similarly,
 $\finally$ is encoded in constraint (\ref{eq:Future_Disc}), where we look for
 the first future position $t'$ where its child has non-negative valuation. The score of its
 child is exponentially scaled so as to diminish the contribution from an
 occurrence far away from the start.
 Finally, we optimize the score of the template using the score of the
root node  $y^\tau _{1, 0}$ w.r.t. the constraint system $\Phi$ (
Algorithm \ref{algo:karak3}). Then we use the resulting model to fill the holes in the template,
obtaining the optimal formula for the chosen depth. By iterating over $\depth$, we may
obtain the minimal such formula.
Next, we establish the correctness and soundness of this algorithm.

\section{Experiments}
\label{sec:expts}
We now present details of our implementation of Algorithm \ref{algo:llm_ltl}, including the design of $\ourtool$, evaluation benchmarks, experimental setup, and experimental results. 
Our tool, $\ourtool$ implements the algorithm in Python and employs $\gurobi$~\cite{gurobi}~as a constraint solver. 
It exposes several configurable hyperparameters to the designer, as shown in Figure~\ref{fig:tool}, including the choice of LLMs (OpenAI's $\chatgpt$~\cite{brockman2016openai}, Anthropic's $\claude$~ and, Google's $\bard$), prompting strategy, templates, and semantics. 

\usetikzlibrary{arrows.meta, positioning, shapes.geometric}

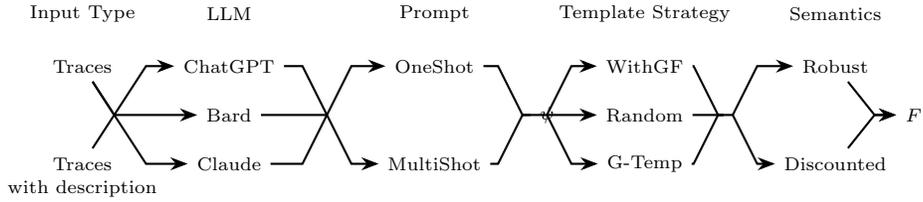
\begin{figure}[t!]
\begin{tikzpicture}[>=Stealth, thick,scale =.65,every node/.style={font=\scriptsize}]

\node (Traces) at (3,0) {Traces};
\node (Traces+Eng) at (3,-2) {Traces};
\node (Traces+Eng+below) at (3,-2.5) {with description};
\node[above of=Traces+Eng] (TracesType) {Input Type};
\node (ChatGPT) at (6,0) {ChatGPT};
\node (Bard) at (6,-1) {Bard};
\node (Claude) at (6,-2) {Claude};
\node[above of=Claude] (LLM) {LLM};
\node (Prompt) at (10.2,0) {OneShot};
\node (Least Pmpt) at (10.2,-2) {MultiShot};
\node[above of=Least Pmpt] (promptType) {Prompt};
\node (Formula) at (12.5,-1) {$\psi$};
\node (WithGF) at (14.5,0) {WithGF};
\node (Random) at (14.5,-1) {Random};
\node (G-Temp) at (14.5,-2) {G-Temp};
\node[above of=G-Temp] (tempType) {Template Strategy};
\node (Template) at (16.3,-1) {};

\node (Sem1) at (18.4,0) {Robust};
\node (Sem2) at (18.4,-2) {Discounted};
\node[above of=Sem2] (semantics) {Semantics};
\node (Finalformula) at (20,-1) {$F$};

\draw[->](Traces) -- ++(0.65,-1) -- (Bard);
\draw[->](Traces+Eng) -- ++(1.3,2) -- (ChatGPT);
\draw[->](Traces) -- ++(1.3,-2) -- (Claude);

\draw[->](ChatGPT) -- ++(1.5,0) -- ++(.5,-1) -- ++(.5,+1) -- (Prompt);
\draw[-](Bard) -- ++(2,0) ;
\draw[->](Claude) -- ++(1.5,0) -- ++(.5,1) -- ++(.5,-1) -- (Least Pmpt);

\draw[-](Prompt)  -- ++(1.3,0)-- ++(.5,-1)  -- (Formula);
\draw[-](Least Pmpt)  -- ++(1.3,0)-- ++(.5,1)   -- (Formula.center);

\draw[->](Formula)  -- ++(0,0)-- ++(.5,1)   -- (WithGF);
\draw[->](Formula.center)  -- ++(0,0) -- (Random);
\draw[->](Formula)  -- ++(0,0)-- ++(.5,-1)   -- (G-Temp);

\draw[-](WithGF)  -- ++(1,0)-- ++(.5,-1)   -- (Template.center);
\draw[-](Random)   -- (Template);
\draw[-](G-Temp)  -- ++(1,0)-- ++(.5,1)   -- (Template);
\draw[->](Template.center)  -- ++(0,0)-- ++(.5,1)   -- (Sem1);
\draw[->](Template)  -- ++(0,0)-- ++(.5,-1)   -- (Sem2);
\draw[->](Sem1)  -- ++(.8,-1) -- (Finalformula);
\draw[->](Sem2)  -- ++(.8,1) -- (Finalformula);
 
\end{tikzpicture}
\caption{The high-level architecture of $\ourtool$.}
\label{fig:tool}
\end{figure}

\subsection{LLM Setup} 
To inform the LLM regarding the syntax and semantics of our specifications and demonstrations, we provide a fixed prompt to the LLM with each example (see, Appendix~\ref{sec:chats}).
This prompt also includes the set of salient predicates present in traces.

For each experiment, we also supply the LLM with the traces and explanation (natural language description) of the desired formula and ask the LLM to generate five LTL formulas such that each formula should satisfy all traces and resemble the English description.
We also conduct an ablation study, where we repeat the test without the explanation.
$\ourtool$ verified each LLM generated formula on the rules previously given; if any failed, we requested a new formula.
Due to cost limitations, we used the web interface to interact with $\chatgpt$ and the other LLMs, thereby limiting the size and quantity of our experiments. As an example of the limitations, the free version of Claude has a strict rate limit restricted the number of experiments that could be performed in a 24 hour period and a size limit on the input.

\begin{table}[b!]
\small
    \centering
    \caption{Example LTL formulas with their natural language description and source.}
    \scriptsize
    \input{formula-list}
    \label{tab:LTL_formula}
\end{table}

\begin{table}[htbp]
    \centering    \caption{$\ourtool$~\semanticsone performance over LLM with description. We show the LLM ouput used for template generation. We present the  formula with maximum manual score, the sum of their fitness over all the input traces (fit columns), and if the formula satisfies all traces. The times are in seconds.}   
    \label{tab:With_Des_one}
    \scriptsize
\input{semantics-results-trace-description-sem1}
\end{table}

\begin{table}[htbp]
    \centering    \caption{$\ourtool$~\semanticstwo performance over LLM with Description. For a detailed description of columns look at caption of Table~\ref{tab:With_Des_one}}  
    \label{tab:With_Des_two}
\scriptsize
    \input{semantics-results-trace-description-sem2}
\end{table}

\begin{table}[htbp]
    \centering    \caption{Performance of LLM with and without traces.}  
    \label{tab:Without_trace}
\scriptsize
    \input{semantics-results-description}
\end{table}

\subsection{Experimental Results}

We synthesize templates from the LLM-produced formula with the best fitness using three different strategies. In the first strategy, we implemented a general randomization strategy, which involved modifying the tree structure by selecting random nodes and replacing them with a tree of height $\leq d$ as mentioned in Section 4. The second entailed the specific fixation of the operators $\globally$ and $\finally$ in addition to the first. This adjustment was integrated as a direct response to the LLM's tendency to generate outputs that closely correspond with the English descriptions but lacking the necessary temporal operators. The third strategy focused on adding random quantifiers at the root node, a measure to maintain the inherent structure of the formulas generated by the LLM, and then synthesizing the template by the first strategy. Among these three strategies, the second yielded the least effective results, contributing minimally to enhancing formula fitness. Conversely, the third strategy proved to be beneficial in elevating formula fitness. Additionally, the randomization approach demonstrated superior performance.
In our experiments, we used mixed of the above strategies and repeatedly ran the tool until it yielded formulas with high fitness.

$\ourtool$ implements formula synthesis using both of our described semantics using fitness-based encoding for both of the semantics to find the best formula using $\gurobi$.
We add additional constraints to ensure that $\gurobi$ did not return a trivial formula by limiting its search space, e.g., formulas like $p\lor\lnot p$ (tautology) or $FFp$ or equivalent sub-formulas like $F(p) \land F(p)$ and various other optimization strategies to reduce the search space.
We ran $\gurobi$ with a time limit and if the time limit was hit, it returned the best solution found so far. 

Table \ref{tab:LTL_formula} shows a small selection of the seventeen formulas used in our experiments as examples of the original formula and natural language description. We originally looked at the formulas presented in~\cite{DBLP:conf/atal/AfzalGGS0V23}, 
but the LLM could solve them without any issues, which answers our first research question 1 positively.
To exercise our tool we required more complicated formulas or a conjunction of many smaller formulas (such as our motivating example).

Each of our experiment formulas required a set of traces and a natural language description.
To obtain such inputs, we started with an LTL formula which was converted to a B{\"u}chi automaton using 
Spot\cite{duret2016spot}, a platform for LTL and $\omega$ -automaton manipulation. We non-deterministically walked the resulting automaton to generate a set of traces that each ended in an accepting states. The natural language description for the formula was written manually creating a scenario that reflected the formula logic.

Tables~\ref{tab:With_Des_one} and~\ref{tab:With_Des_two} present results of our ~\semanticsone and \semanticstwo experiments for the formulas in Table \ref{tab:LTL_formula}. In these tables, we show the resulting best formula obtained by $\ourtool$ after evaluations. Final formula evaluation was done three ways: fitness score, satisfaction with input traces, and manual grading. Manual grading on a scale of 1 to 5 was based on the extracted formula's expression of the original formula's safety and liveliness requirements along with how well it met the intent of the overall specification. For each experiment we've recorded the run time of $\ourtool$ alone because LLM response was typically short, benefiting from large cloud computing resources, while the optimization query in $\ourtool$ runs significantly longer on a local PC. 

In the end, 16 of the 17 the LLM generated and tool corrected formulas were sufficient to express the intent of the original formulas. In the case of the failing formula, the LLM was unable to generate any useful response. Additionally, we found that $\ourtool$ has improved the quality for 12 of the LLM generated formula.

In our experiments, LLM produced formulas with better quality for 10 and the same quality for 3 inputs when traces were provided.

Once we introduce the natural description of the original formula LLM performance improved significantly, which answers our research question 2 positively.

We observe for most of the cases, an LLM can find formulas that are reasonably close to the description and $\ourtool$ can improve the quality of the resulting formulas most of the time with good fitness and similarity to the original formula as manually judged. 

We found that both semantics produced interesting formulas though \semanticstwo had a higher incidence of producing formulas that are not satisfied by all the traces.

\subsection{Ablation Study}
As an ablation study, we tested the impact of the natural language description in the LLM's performance by providing only traces for the formula generation step of our process. We found that the LLMs struggled to produce any useful formula. Due to this poor performance we did not include them in our experimental results. We repeated the ablation testing by removing the traces from the process to examine the impact of the descriptions alone. Table~\ref{tab:Without_trace} contrasts the learned formulas and their scores for the cases when we provided traces and when we did not. We observed that the LLM's success was dependent on several factors, the given detail of the original descriptions, their degree of specificity, and the LLM's ability to understand how the temporal operators F and G define acceptance of future parts of the trace. Finally, to test the contribution of the LLM and $\ourtool$ to the overall results, $\ourtool$ was given the formula traces and an empty template, testing if it could generate the formulas from scratch. Within the given processing timeout, no useful formulas were obtained. This result answers research question 3 positively.  

We trained the LLM models in two modes: one-shot learning and multishot learning. In one shot learning, we trained the LLM using examples and ground rules to give the LLM an understanding of the LTL and the synthesis problem. The chat is presented in Appendix \ref{sec:chats}. In multishot learning, we divide the problem into sub-tasks and ask the LLM to generate a formula.  Initially, we provided the English description plus one trace. Subsequently, we increased the subset size to three traces and then to six traces. After each increase in traces, we ask the LLM to analyze and verify the given traces alongside the English description. We observed the LLM give an average formula at the first step, and with each addition of traces, it keeps adding small modifications to the formula using LTL operators, which results in poor performance. We found the first method more effective than the latter one.
Our reported results are using one-shot learning with $\chatgpt$. 

\section{Related Work}
\label{sec:related}

Grammatical inference~\cite{lopes2009active} is concerned with learning formal languages as grammars and their automata representations \cite{grammaticalInference,grammaticalInferenceBook}. Active techniques rely on querying the system under learning to guide the inference process, whereas passive grammatical inference leverages a static set of trace behavior without making further queries for additional data. The former is exemplified by $L^\star$ algorithm \cite{angluin1987learning}, whereas the latter generally relies on state-merging procedures and can be used to learn probabilistic automata \cite{ALERGIA}, MDPs in the context of model checking \cite{IO_ALERGIA_1}, timed automata \cite{IO_ALERGIA_2}, and regular decision processes \cite{AbadiBrafman20}.

The problem of learning LTL formulae from traces is a form of grammatical inference that has been well-studied. Several of the methods most related to our own are presented in \cite{lemieux2015general,ltlFMCAD18,roy2023learning,vazquez2018learning}. The focus of \cite{ltlFMCAD18} is to produce the minimal formula which is consistent with a rational sample represented as a lasso. The problem of matching formulae with traces is encoded as a constraint system and a satisfying assignment yields the learned property. However, requiring the inputs to be lassos significantly restricts the application to real scenarios. The method in \cite{lemieux2015general} requires a user-defined input template of the LTL formula which they would like to satisfy and outputs all possible propositional substitutions consistent with the sample. Kim \cite{kim2019bayesian} select the best candidate LTL templates that satisfy the given positive and negative traces based on Bayesian probability.

Unlike these works, we use positive traces generated without restrictions for our learning. While \cite{gold1967language} has shown the difficulty of this approach, recent work has focused on method to overcome the restriction. Roy \cite{roy2023learning} explored a method that progressively builds a DFA from a trivial single satisfying state to the smallest non-trivial DFA that still maintains satisfaction of the provided traces.  Another form of progressively building a to final formula was explored in \cite{vazquez2018learning}. Their approach started with a known good formula and a set of demonstrations that defined a new, related specification. Using a trellis structure, they progressively update the formula, pruning the sections that do not extend the formula in support of the additional specification. 
Our approach works in a progressive manner but, unlike these approaches, we start with candidate LLM generated LTL formulas that are expected to contain errors and uses small templates of LTL syntax to improve the overall satisfaction of formulas against positive traces.

Extracting formal requirements from natural language text has been a focus of extensive research \cite{brunello2019synthesis}. While there has been progress using unrestricted natural language  specifications \cite{ilieva2005automatic,ghosh2016arsenal}, many techniques require conformance of the requirements language to specific ontologies \cite{sadoun2013natural}, sentence structure \cite{yan2015formal} or unique requirements languages \cite{giannakopoulou2020generation,carvalho2015nat2test,santos2018formal}.
Recently \cite{liu2023lang2ltl,DBLP:journals/corr/abs-2206-01962,rahmani2021multi,cosler2023nl2spec}, large language models have been utilized to translate LTL from natural language.
Liu et al. \cite{liu2023lang2ltl} propose Lang2LTL, a modular system that leverages pretrained LLMs to extract referring expressions, corresponding real-world landmarks, and finally translate the command into an LTL task specification for robotic motion planning.
Hahn et al. \cite{DBLP:journals/corr/abs-2206-01962} study generalization abilities of LLMs when translating natural language into formal specifications with complex semantics.
They report that LLMs maintain their generalization capabilities from pre-trained knowledge of natural language to generalize to variable names and operator descriptions. 
However none of these approaches study the application of LLMs in extracting LTL from demonstrations or descriptions. 
Mavrogiannis \cite{mavrogiannis2023cook2ltl} uses an LLM trained with task examples and allowed instructions for translating cooking instructions from random internet recipes into LTL formulas. 
Closer to our approach is that of Rahmani et al. \cite{rahmani2021multi} where the authors combine LLMs to generate candidate formulas directly from a task description and then use component-based synthesis to improve on the results.
In contrast, our work is focused on using an LLM not to generate exact LTL formulas but synthesizing candidate ones from positive, finite traces that could then be repaired to maximize the best qualities of both the LLM and existing LTL synthesis tools.

\bibliographystyle{IEEEtranS}
\bibliography{papers}

\newpage 
\appendix

\section{Rationale behind Robustness Semantics}
In the robust semantics \cite{DBLP:conf/atal/AfzalGGS0V23}, given an LTL formula $\varphi$ and a trace $T$, a 
 valuation function  $\sem{\varphi, T}$ is defined as follows. 
\begin{itemize}
\small
    \item
    $ \sem{(p,T)} = 
    \begin{cases} 
    1 & \text{if } p \in T(0), \\
    -1 & \text{if } p \notin T(0).
    \end{cases}
    $
    \item 
    $
    \sem{(\neg p,T)} = 
    \begin{cases} 
    1 & \text{if } p \notin T(0), \\
    -1 & \text{if } p \in T(0).
    \end{cases}
    $
\small
    \item 
    $
    \sem{(\varphi \land \psi, T)} =
    \begin{cases}
        \beta \cdot \sem{(\varphi, T)} \cdot \sem{(\psi, T)}   & \text{if~}  \sem{(\varphi, T)} \geq 0 \land \sem{(\psi, T)} \geq 0, \\
        -1 & \text{otherwise.}
    \end{cases}
    $
    \item
    $
    \sem{(\varphi \lor \psi, T)} =  \beta \cdot
    \begin{cases}
     \frac{(\sem{(\varphi, T)}+ \sem{(\psi, T)})}{2} & \text{if~}  \sem{(\varphi, T)} \geq 0 \wedge  \sem{(\psi, T)} \geq 0,\\
 \max(\sem{(\varphi, T)}, \sem{(\psi, T)}) &  \text{otherwise.}   
    \end{cases}
    $
\small
    \item %Globally ($G$): 
    $
    \sem{(G\varphi, T)} = \beta \cdot
    \begin{cases} 
     \displaystyle\sum_{i=0}^{|T|} \alpha^i \sem{(\varphi, T(i))} & \text{if } \sem{(\neg\varphi, T(t))}< 0 \hspace{0.1cm}\forall t, \\
    -1 & \text{otherwise}.
\end{cases}
    $
    \item %Finally ($F$): 
    $
    \sem{(F\varphi, T)} = \beta \cdot
    \begin{cases} 
     \alpha^t \sem{(\varphi, T(t))} & \text{where } t = \min\{j | \sem{(\varphi, T(j))} \geq 0\}, \\
    \gamma*\alpha^{|T|} & \text{if } \sem{(\varphi, T(t))} = 0 ~\text{for all}~ t.
\end{cases}
    $
    \item 
    $
    \sem{(X\varphi, T)} = 
    \begin{cases} 
          \sem{(\varphi, T(1))} & \text{if } \sem{(\varphi, T(1))}\geq0  
          \land |T|>1  \\
          \gamma & \text{if~} |T|<2 \\
        -1 & \text{otherwise}.
    \end{cases}
    $
    \item
    $
    % V(X\varphi, T) = \beta \cdot V(\varphi, T(2)).
    \sem{(\varphi U \psi, T)} = 
    \begin{cases} 
       \alpha^t \cdot \sem{(\psi, T(j))} & \text{where}~ t=\min\{j | \sem{(\psi, T(j))} \geq 0\}
      \text{~and}~\\
      & \forall i < t,~ \sem{(\varphi, T(i))} \geq 0  \\
      \alpha^{|T|} \cdot \gamma  & \text{if } \forall i \sem{(\psi, T(i))} \geq 0 \\
      
     -1 & \text{ otherwise}.
\end{cases}
    $
\end{itemize}
For the literals, we assign valuation -1  or 1 depending on whether 
the trace satisfies it or not. We interpret conjunction by taking the product of the valuation of the individual conjuncts. This  implies that both subformulae must do  well on the trace. This is better than assigning $\sem{(\varphi {\land} \psi, T)}$ as the minimum of $\sem{(\varphi, T)}$ and $\sem{(\psi, T)}$~\cite{tabuada2015robust}. This valuation ensures that both $\varphi$ and $\psi$ must score high for $\varphi \land \psi$ to score high. However, the $\min$ function is not sensitive to the formula that has a higher value. Therefore, the learning algorithm becomes unguided for one part of the formula. This suggests a modification to the valuation function that takes both the subformulae into account symmetrically, without flattening one of the subformulae. Our valuation function for the conjunction, defined as their product $\sem{(\varphi, T)} \cdot \sem{(\psi, T)}$, is based on this idea. 
 
We interpret disjunction as an average over the disjuncts when the valuations of individual disjuncts 
are $\geq 0$.  Likewise, if one of the disjuncts is $\geq 0$ and the other is $< 0$, then we choose the maximum value. 
This  implies that a disjunction scores well 
if any of the two disjuncts does well on the trace. 

Our interpretation of $\globally \varphi$ computes the discounted sum of the value of $\varphi$
at each position of the word. To give less weightage to a literal appearing further down in the trace as opposed to earlier, 
 we apply the temporal discount of $0 < \arrr < 1$.
Our interpretation of $\finally \varphi$ computes the discounted score of $\varphi$
at the earliest position where $\varphi$ has a non-negative score.

The discount value  $\arrr$ is a parameter defined by the user to control the temporal complexity.
Consider the
formula $\varphi=\finally q$ for $p, q \in \prop$ and the trace
$$T=\{p\}\{p,r\}\{p\}\{p,s\}\{p\}\{p\}\{p,q\}(\{r\}\{q\})^*$$
Then $q$ holds for
the first time at $T(7)$. Thus, $V(q,T[7:|T|])=1$, making $V(\varphi, T)=\arrr^{7}$.
Our valuation assigns non-zero scores only for satisfiable formulae.

\section{Rationale for Discounted Semantics}
Let $0 {\leq} \alpha {\leq} 1$ and $0 {\leq} \beta {\leq} 1$ be temporal and nesting discount factors.
For a finite trace $w$ and LTL formula $\varphi$, we define the valuation of $\varphi$ wrt $w$,  $\sem{\varphi, w}$ as follows:
\begin{itemize}
    \item $\sem{\true, w} = 1$.
    \item $\sem{a, w} = 1$ if and only if $w(0) \ni a$.
    \item $\sem{\neg \varphi, w} = 1 - \sem{\varphi, w}$.
    \item $\sem{\varphi \vee \psi, w} = \beta \cdot \max \set{\sem{\varphi, w}, \sem{\psi, w}}$.
    \item $\sem{\Next \varphi, w} = \alpha \sem{\varphi, w[1]}$.
    \item $\sem{\varphi \until \psi, w} = \max\limits_{0 \leq i < |w|} \set{\min \set{\alpha^i \cdot \sem{\psi, w[i]}, \min\limits_{0 \leq j < i} \set{\alpha^j \cdot \sem{\varphi, w[i]}}}}.$
\end{itemize}
For $\globally$ and $\eventually$ operators, these semantics follow from those of $\until$ and given here for completeness: 
\begin{itemize}
    \item $\sem{\eventually \varphi, w} = \max\limits_{0 \leq i < |w|} \set{\alpha^i \cdot \sem{\varphi, w[i]}}.$
   \item 
   $\sem{\globally \varphi, w} = 1 - \max\limits_{0 \leq i < |w|} \set{\alpha^i \cdot \sem{\neg \varphi, w[i]}} = 1 - \max\limits_{0 \leq i < |w|} \set{\alpha^i \cdot ( 1 - \sem{\varphi, w[i]})} .$
\end{itemize}

For a trace $w = \seq{\set{\neg p}\set{\neg p}\cdots\set{\neg p}\set{p}\set{\neg p}\set{\neg p}}$, $\sem{\eventually p, w}$ evaluates to $\alpha^n$ where $n$ is the first index at which $p$ becomes true.
Similarly, for a trace $w = \seq{\set{p}\set{p}\cdots\set{ p}\set{\neg p}\set{p}\set{p}}$, $\sem{\globally p, w}$ evaluates to $1 - \alpha^n$ where $n$ is the first index in which $p$ becomes false. 
In summary, $\sem{\eventually p, w}$ is higher when $p$ is found earlier, capturing the quantitative semantics for ``eventually''. 
On the other hand, $\sem{\globally p, w}$ is higher for a longer contiguous initial sequence that the trace exhibits, capturing the quantitative semantics of ``globally.''

\section{Trace Constraints for the Robust Semantics}
\label{app:rob-cons}
 To encode the formula structure according to the robust semantics,
  we use a syntax tree with  identifiers
$\nodes=\{1, 2, \ldots, n\}$ for the set of nodes, where $n = |\nodes|$.
We assume herein that the root node is
identified as $1$.  We have the child relations $L$ and $R$, such that $(i, j)
\in L$ (or $R$) iff the node $j$ is the left (right) child of node $i$. The only child of unary operators  
is considered as a  left child by assumption. For each node $i \in \nodes$ and
possible label $\lambda \in \op \;\union\; \prop$, we introduce a Boolean variable
$x_{i, \lambda}$ indicating whether the node is labelled with an operator
($\op$) or variable ($\prop$).

\noindent{\bf{Trace Constraints for the Robust semantics}}: We describe the functional constraints  imposed by the operators. For this, to each node $i \in \nodes$, we attach a set of real variables $Y^\tau _i = \{y^\tau _{i, t} \; \mid 0 \leq t \leq |\tau|, i \in \nodes\}$ representing its
 \emph{score} at each point $0 \leq t \leq |\tau|$ in a trace $\tau \in  T$, where $y^\tau _{i, t}$ is defined below.  
 For a given trace $\tau$ and for \semanticsone we construct $\varphi^\tau _\depth$ as the conjunction of:
 {\footnotesize
\begin{gather}
  y^\tau _{1,0} >= 1 \label{eq:pos_def} \\
  %\textstyle
    \bigwedge\limits_{1 \leq i \leq n} \bigwedge\limits_{p \in \prop }\;\;x_{i, p} \Rightarrow \left[ 
        \bigwedge\limits _{1 \leq t \leq |T|} y^\tau _{i, t} = \begin{cases}
             1 & ~if~ p \in \tau (t) \\
            -1 & ~if~ p \not \in \tau (t) 
        \end{cases}
      \right] \label{eq:label_score} \\
      \bigwedge\limits_{1 \leq i \leq n} \bigwedge\limits_{p \in \prop }\;\;x_{i, !p} \Rightarrow \left[ 
        \bigwedge\limits _{1 \leq t \leq |T|} y^\tau _{i, t} = \begin{cases}
             -1 & ~if~ p \in \tau (t) \\
            1 & ~if~ p \not \in \tau (t) 
        \end{cases}
      \right] \label{eq:not_l_score} \\
    \bigwedge\limits _{\substack{1 \leq i \leq n, \\(i, j) \in L, \\(i, j') \in R}}\;\;x_{i, \land} \Rightarrow \left[ 
        \bigwedge\limits _{1 \leq t \leq |T|} y^\tau _{i, t} = \begin{cases}
         \beta \cdot y^\tau _{j,t}\cdot y^\tau _{j', t} & ~if~  y^\tau _{j,t}\geq0\land y^\tau _{j', t}\geq0 \\
         -1 & \textnormal{otherwise}
        \end{cases} 
      \right] \label{eq:and_score} \\
    \bigwedge\limits _{\substack{1 \leq i \leq n, \\(i, j) \in L, \\(i, j') \in R}}\;\;x_{i, \lor} \Rightarrow \left[        \bigwedge\limits _{1 \leq t \leq |T|}  y^\tau _{i, t} = \beta \cdot \begin{cases}
        \frac{y^\tau _{j,t} + y^\tau _{j',t}}{2} & y^\tau _{j,t}\geq0\land y^\tau _{j',t}\geq0 \\
       \max\limits _{} \{  y^\tau _{j,t} , y^\tau _{j',t} \} & otherwise \end{cases}
      \right] \label{eq:or_score}  \\
       \bigwedge\limits _{\substack{1 \leq i \leq n, \\(i, j) \in L, \\(i, j') \in R}}\;\;x_{i, \implies} \Rightarrow \left[         \bigwedge\limits _{1 \leq t \leq |T|} y^\tau _{i,t} =\beta \cdot \begin{cases}        \frac{-y^\tau _{j,t} + y^\tau _{j',t}}{2} & y^\tau _{j,t}<0\land y^\tau _{j',t}\geq0 \\
        \max\limits _{} \{ -y^\tau _{j,t} ,   y^\tau _{j',t} \} & otherwise
      \end{cases} 
      \right] \label{eq:implies_score} \\    
    \bigwedge\limits _{\substack{1 \leq i \leq n, \\(i, j) \in L}}\;\;x_{i, G} \Rightarrow \left[ 
       \bigwedge\limits _{1 \leq t \leq |T|} y^\tau _{i,t} = \beta \cdot
       \begin{cases}    
       \sum^{|T|}_{t'=t} \alpha^{t'} \cdot y^\tau _{j,t'},& where ~\forall t', y^\tau _{j,t'}\geq 0 \\
       -1 & otherwise 
       \end{cases}
      \right] \label{eq:Globally} \\
    \bigwedge\limits _{\substack{1 \leq i \leq n, \\(i, j) \in L}}\;\;x_{i, F} \Rightarrow \left[ 
         \bigwedge\limits _{1 \leq t \leq |T|} y^\tau _{i,t} = \beta \cdot
         \begin{cases}
         \alpha^{t'}  y^\tau _{j,t'} ,& \textnormal{where }  t\leq t'< |T|,t' = \min\{i \geq t | y^\tau _{j,i}\geq0\} \\
         \gamma \cdot \alpha^{|\tau|} & otherwise
         \end{cases}
      \right] \label{eq:Future} \\
    \bigwedge\limits _{\substack{1 \leq i \leq n, \\(i, j) \in L}}\;\;x_{i, X} \Rightarrow \left[       \bigwedge\limits _{1 \leq t \leq |T|}  y^\tau _{i,t} = 
      \begin{cases}
      y^\tau _{j,t+1}, & \textnormal{where, } y^\tau _{j,t+1} \geq0 \\
      \gamma & ~if~ t+1 \geq |\tau|  \\
      -1 & otherwise
      \end{cases}
      \right] \label{eq:next} \\
    \bigwedge\limits _{\substack{1 \leq i \leq n}}\;\;x_{i, U} \Rightarrow \left[         \bigwedge\limits _{1 \leq t \leq |T|}   y^\tau _{i,t} = \beta \cdot \begin{cases}
              \alpha^{t''}\cdot y^\tau _{j,t''} & ~if~ \forall t', t\leq t' < t'',y^\tau _{j,t'}\geq0\land y^\tau _{j,t''} \geq 0  \\
              \gamma \cdot \alpha^{|\tau|} & ~if~ \forall i y^\tau _{j,t'}\geq0 \\
            -1 & \textnormal{  otherwise}
        \end{cases}
      \right] \label{eq:Untill} 
\end{gather}
}%

Corresponding to Boolean variables and to each operator, the constraints encode
the calculation of the valuation of a given node as a function of the valuation
of its children, as defined in Section \ref{sec:robust}.

\section{Information about LLM (ChatGPT) Prompts}
\label{sec:chats}

% \section{Chats with $\chatgpt$}

% \vspace{-3mm}
\vbox{
$\text{ChatGPT}$ chat defining the rules the LLM must use for formula generation:
\vspace{-3mm}
\begin{center}
\fbox{
\begin{minipage}{0.95\textwidth}
\small
\textbf{Training:} This prompt aims to generate LTL formulas by observing a pattern in given traces and taking hints from English description.
English description is like a hint that gives us an overview of what is happening in traces.

\textbf{ChatGPT:} Sure! For those unfamiliar, LTL stands for "Linear Temporal Logic". It's a type of logic used to describe properties of sequences of states in time. LTL formulas are commonly used in the domain of formal verification, especially for describing and checking the behaviors of systems over time.
To generate an LTL formula based on provided traces and English descriptions, I'll need:

1. The traces, which are sequences of states or events that you've observed.

2. The English description that serves as a hint.

Once you provide both, I'll do my best to capture the observed behavior as an LTL formula. Please go ahead and provide the required details.
\end{minipage}
}
\end{center}

\begin{center}
\fbox{
\begin{minipage}{0.99\textwidth}
\small
\textbf{Training:} These are the definitions of LTL terms:
\begin{enumerate}
    \item $\pi \models \top$
    \item $\pi \not\models \bot$
    \item $\pi \models p$ iff $p \in L(s_1)$
    \item $\pi \models \neg \phi$ iff $\pi \not\models \phi$
    \item $\pi \models \phi_1 \land \phi_2$ iff $\pi \models \phi_1$ and $\pi \models \phi_2$
    \item $\pi \models \phi_1 \lor \phi_2$ iff $\pi \models \phi_1$ or $\pi \models \phi_2$
    \item $\pi \models \phi_1 \rightarrow \phi_2$ iff $\pi \models \phi_2$ whenever $\pi \models \phi_1$
    \item $\pi \models X \phi$ iff $\pi_2 \models \phi$
    \item $\pi \models G \phi$ iff, for all $i \geq 1$, $\pi \models \phi$
    \item $\pi \models F \phi$ iff there is some $i \geq 1$ such that $\pi_i \models \phi$
    \item $\pi \models \phi U \psi$ iff there is some $i \geq 1$ such that $\pi_i \models \psi$ and for all $j = 1, \ldots, i - 1$, we have $\pi_j \models \phi$
\end{enumerate}

$\phi$ is any formula:

$\phi ::= \bot \,|\, p \,|\, (\neg \phi) \,|\, (\phi \land \phi) \,|\, (\phi \lor \phi) \,|\, (\phi \rightarrow \phi) \,|\, (X \phi) \,|\, (F \phi) \,|\, (G \phi) \,|\, (\phi U \phi)$

\textbf{ChatGPT:} Understood. This is a basic definition of LTL semantics. You've presented the meaning of various LTL operators in terms of how they relate to traces (denoted as $\pi$).

To make the task clear:

You will provide some traces (sequences of states/events).
You will provide an English description hinting at the behavior captured in those traces.
Based on the traces and the English description, I'll generate an LTL formula capturing the observed behavior.

Please provide the traces and the English description so that I can generate the LTL formula.
\end{minipage}
}
\end{center}
} %Vbox

\begin{center}
\fbox{
\begin{minipage}{0.9\textwidth}
\label{RulesforLTL}
\small
\textbf{Training:} Rules to follow for generating formulas:

\textbf{Note -}
\begin{enumerate}
    \item The formula should include all variables present in traces.
    \item If there are '1's at the end of any trace, that means there is no pattern in traces. There is no pattern. So, there cannot be 'G' in the beginning of the formula.
    \item If you see terms like globally, for all, always, every time, and other similar terms, use 'G'.
    \item If you see terms like future, at some point, eventually, and other similar terms, try to use 'F'.
    \item If terms like until, till, and similar terms are used, then try to use 'U'.
\end{enumerate}

\textbf{Steps to follow when generating formulas:}
\begin{enumerate}
    \item Follow all the above-mentioned rules carefully.
    \item First, observe and find patterns in traces without even looking at the English description.
    \item Generate 5 LTL formulas using pattern, observation, and English description, such that each formula should satisfy all traces and resemble the English description.
    \item After generating formulas, verify that they follow all the rules; otherwise, regenerate them.
\end{enumerate}

\textbf{ChatGPT:} Understood. I'll follow these steps and rules when generating the LTL formula.

Please provide the traces and the English description, and I'll proceed to generate the LTL formulas based on your specifications.
\end{minipage}
}
\end{center}

\vbox{
LLM Formula Extraction Examples
\begin{center}
\fbox{
\begin{minipage}{\textwidth}
\small
\textbf{Training: Example 1 - Go through each step}

Input Traces where "," is the label separator, ";" is the state separator, and "\#" is the end of the trace:

\texttt{p,!q,!r;!p,!q,!r;!p,!q,!r;p,q,!r;1;1;1;1;1;1\#}\\
\texttt{p,q,r;1;1;1;1;1;1;1;1;1;1;1;1;1;1;1;1;1;1;1;1\#}\\
\texttt{!p,!q,!r;p,q,!r;1;1;1;1;1;1;1;1;1;1;1;1;1;1;1;1;1;1\#}\\
\texttt{p,!q,!r;!p,!q,!r;p,q,!r;1;1;1;1;1;1;1;1;1;1;1;1;1;1;1;1;1;1;1;1;1;\\1;1;1;1;1;1;1;1;1;1;1;1;1;1;1;1;1;1;1;1;1\#}\\
\texttt{p,q,r;1;1;1;1;1;1;1;1;1;1;1;1;1;1;1;1;1;1;1;1\#}\\
\texttt{p,q,!r;1;1;1;1;1;1;1;1;1;1;1;1;1;1;1;1;1;1;1;1\#}\\
\texttt{!p,q,r;p,q,r;1;1;1;1;1;1;1;1;1;1;1;1;1;1;1;1;1;1;1;1\#}\\
\texttt{p,q,r;1;1;1;1;1;1;1;1;1;1;1;1;1;1;1;1;1;1;1;1\#}\\
\texttt{p,!q,!r;p,q,r;1;1;1;1;1;1;1;1;1;1;1;1;1;1;1;1;1;1;1\#}\\
\texttt{p,q,!r;1;1;1;1;1;1;1;1;1;1;1;1;1;1;1;1;1;1;1;1\#}\\
\texttt{1;1;1;1;1;1;1;1;1;1;1;1;1;1;1;1;1;1;1;1;1;1\#}

\textbf{English Description:} "At some point during its operation, the software will experience a critical error (\texttt{p}) and will either generate an error report (\texttt{q}) or initiate a self-repair process (\texttt{r})."

\textbf{ChatGPT:} Let's break down the task step by step, as per your instructions:

1. Observe and find patterns in traces:
Looking at the traces, some consistent patterns that emerge:

- There are traces with the sequence: \texttt{!p,!q,!r;} followed by \texttt{p,q,!r;} and then a series of \texttt{1;} states.
- There are traces with the sequence: \texttt{p,q,r;} followed by a series of \texttt{1;} states.
- Some traces have only \texttt{1;} states.
From this, we can deduce a few patterns:

- The critical error \texttt{p} happens either alone or with other events \texttt{q} or \texttt{r}.
- After the occurrence of \texttt{p}, the sequence tends to end or goes into a series of \texttt{1;} states.

2. Generate 5-LTL formulas:
Given the English Description, we are looking for the occurrence of \texttt{p} and then either \texttt{q} or \texttt{r}.

Considering the provided information and patterns observed:

- $\texttt{F(p \& (q | r))}$
  This formula ensures that at some point, there's a state where \texttt{p} occurs and either \texttt{q} or \texttt{r} happens as well.

- $\texttt{F(p \& q) | F(p \& r)}$
  This formula breaks down the occurrences into two separate futures, one where \texttt{p \& q} happens and another where \texttt{p \& r} happens.

- $\texttt{F(p) -> X(q | r)}$
  This formula captures the essence that after \texttt{p} happens, in the next state either \texttt{q} or \texttt{r} will happen.

- $\texttt{F(p) \& (F(q) | F(r))}$
  This formula ensures that \texttt{p} will happen at some point and also either \texttt{q} or \texttt{r} will happen at some point. This is a bit weaker than the other formulas.

- $\texttt{G((p \& !q \& !r) -> X(q | r))}$
  This formula globally ensures that whenever there's a state where \texttt{p} happens and \texttt{q} and \texttt{r} don't, in the next state either \texttt{q} or \texttt{r} should happen.

3. Verification:
All the above formulas follow the given rules:

- They include all the variables present in traces: \texttt{p}, \texttt{q}, \texttt{r}.
- They do not use \texttt{G} in the beginning as there are series of \texttt{1;} states in the traces, indicating no global pattern.
- They align with the given English description.
- The most relevant formula, as you pointed out, is $\texttt{F(p \& (q | r))}$. This formula aptly captures the essence of the English description.
\end{minipage}
}
\end{center}
} %Vbox

\begin{center}
\fbox{
\begin{minipage}{\textwidth}
\small
\textbf{Training: Example 2 - Go through each step}

Input Traces where “,” is the label separator, “;” is the state separator, and “\#” is the end of trace:

\texttt{!p,q,r,s;!p,q,r,s;!p,!q,!r,!s;p,!q,!r,!s;p,!q,!r,!s;p,!q,r,!s;p,!q,!r,!s;\\!p,q,!r,s;1;1\#}\\
\texttt{!p,q,!r,!s;!p,!q,r,!s;p,q,r,s;!p,q,r,s;1;1;1;1;1;1\#}\\
\texttt{!p,q,r,s;p,q,r,!s;!p,q,r,!s;1;1;1;1;1;1;1\#}\\
\texttt{!p,q,r,s;p,q,!r,!s;!p,q,!r,s;1;1;1;1;1;1;1\#}\\
\texttt{p,q,r,!s;!p,q,!r,!s;p,q,!r,!s;p,q,!r,!s;p,q,r,s;1;1;1;1;1\#}\\
\texttt{!p,q,r,!s;p,q,!r,s;p,q,!r,s;1;1;1;1;1;1;1\#}\\
\texttt{p,!q,r,s;p,q,r,s;1;1;1;1;1;1;1;1\#}\\
\texttt{p,!q,r,!s;p,q,!r,s;1;1;1;1;1;1;1;1\#}\\
\texttt{p,q,r,s;p,q,r,!s;1;1;1;1;1;1;1;1\#}\\
\texttt{!p,!q,!r,s;!p,!q,!r,s;p,!q,r,!s;p,q,r,s;1;1;1;1;1;1\#}

\textbf{English Description:} At some point, the software will be updated (p), and in the next time period, the system will verify its integrity (q) and either restart (r) or notify the user of success (s).

Most relevant formula: $F(p \& X(q \& (r | s)))$

ChatGPT:
Let's follow the steps:

1. Observe and find patterns in traces:
Observing the traces, we can identify a few patterns:

- After p occurs (i.e., the software is updated), we see that q often follows in the next state, indicating verification of integrity.
- Once q occurs, we observe that either r or s can happen, representing the system restart or user notification, respectively.

2. Generate 5-LTL formula:
Given the English Description, after the occurrence of p, in the next state q should occur, and then either r or s should happen.

From this understanding, we can form the following LTL formulas:

- $F(p \& X(q \& (r | s)))$ \\
  This formula ensures that at some future point when the software is updated (p), in the very next time frame, it will verify its integrity (q), and then either restart (r) or notify the user (s).

- $F(p \& X(q)) \& F(q \rightarrow X(r \vert s))$ \\
  This formula breaks down the behavior into two separate future events. First, the software gets updated and verifies its integrity. And when integrity is verified, in the next state, the system either restarts or notifies the user.

- $G(p \rightarrow X(q \& (r \vert s)))$ \\
  This formula ensures that globally, whenever the software updates, in the next state it verifies its integrity and either restarts or notifies the user.

- $F(p) \& F(q \& (r \vert s))$ \\
  This formula is a bit more relaxed. It ensures that at some point in the future, the software will update, and at another point (not necessarily immediately after), it will verify integrity and take action (either restart or notify).

- $G(p \rightarrow X(q)) \& G(q \rightarrow X(r \vert s))$ \\
  This formula breaks down the two-step behavior. It ensures that globally, when the software updates, it will verify its integrity in the next state. Also, once integrity is verified, in the next state, the system either restarts or notifies the user.

3. Verification:
All the above formulas follow the given rules:

- They include all the variables present in traces: p, q, r, s.
- They do not start with G because of the series of 1; states in the traces, indicating no global pattern.
- They align with the provided English description.
- The most apt formula, as you mentioned, is $F(p \& X(q \& (r | s)))$. This formula accurately captures the described behavior.
\end{minipage}
}
\end{center}

\begin{center}
\fbox{
\begin{minipage}{\textwidth}
\small
\textbf{Training: Example 3 - Go through each step}

Input Traces where “,” is the label separator, “;” is the state separator, and “\#” is the end of trace:

\texttt{!p,q,r,s;p,q,r,s;!p,!q,!r,s;!p,!q,r,s;p,q,r,!s;p,q,r,s;p,!q,!r,s;\\!p,!q,!r,!s;!p,q,!r,!s;!p,q,r,!s\#}

\texttt{p,!q,r,!s;!p,q,r,s;p,q,r,s;p,!q,r,s;!p,!q,r,s;p,q,r,!s;!p,!q,!r,s;\\!p,q,r,s;!p,!q,r,!s;p,q,r,s\#}

\texttt{p,q,r,!s;!p,!q,!r,s;p,!q,!r,!s;!p,!q,r,!s;p,!q,!r,s;!p,!q,r,!s;\\p,q,r,s;!p,!q,r,s;!p,q,!r,!s;!p,!q,!r,!s\#}

\texttt{!p,!q,r,!s;p,!q,r,s;p,!q,r,!s;!p,!q,r,!s;p,q,r,!s;p,q,r,s;p,q,r,s;\\!p,!q,r,s;p,q,r,!s;p,q,r,s\#}

\texttt{p,!q,r,s;!p,!q,!r,s;p,q,r,s;!p,!q,r,s;!p,!q,r,s;!p,!q,!r,!s;!p,q,!r,s;\\!p,!q,!r,s;p,!q,!r,s;p,!q,r,!s\#}

\texttt{p,!q,r,!s;p,q,r,s;p,q,r,s;!p,!q,r,s;!p,!q,r,s;p,!q,r,!s;p,!q,!r,s;\\!p,q,!r,s;p,!q,!r,s;p,!q,r,!s\#}

\texttt{!p,!q,r,s;p,q,r,s;p,!q,r,s;!p,!q,r,s;p,q,r,s;!p,!q,!r,s;!p,q,!r,!s;\\!p,q,r,s;!p,!q,r,!s;!p,q,!r,s\#}

\texttt{p,!q,r,s;!p,!q,r,s;p,q,r,!s;p,q,r,s;!p,!q,r,s;!p,q,r,!s;p,!q,!r,!s;\\p,!q,!r,!s;p,!q,!r,s;p,q,r,s\#}

\textbf{English Description:} Always, when the room is occupied (p) and the lights are off (q), the motion sensors will activate (r), and in the next time period, the emergency lights will turn on (s).

Most relevant formula:  $G((p\land q)\to(r\land X(s)))$
\end{minipage}
}
\end{center}

\begin{center}
\fbox{
\begin{minipage}{\textwidth}
\small

ChatGPT:
Let's analyze the given traces based on the steps:

1. Observe and find patterns in traces:
From the traces, we can observe:

- The sequence p,q,!r,!s appears several times, indicating that sometimes the room is occupied, the lights are off, but the motion sensors are not activated, and the emergency lights don't turn on.
- The sequence p,q,r,s appears several times, indicating that sometimes when the room is occupied and the lights are off, the motion sensors are activated, and the emergency lights are turned on immediately.

2. Generate 5-LTL formula:
Given the English Description, always when the room is occupied (p) and the lights are off (q), the motion sensors will activate (r), and in the next time period, the emergency lights will turn on (s).

We can generate the following LTL formulas:

- $G((p\land q)\to(r\land X(s)))$ \\
  This formula, which you identified as the most relevant one, captures the behavior that globally, whenever the room is occupied and the lights are off, the motion sensors will activate, and in the next time frame, the emergency lights will turn on.

- $G(p\land q\to X(r\land s))$ \\
  This formula suggests that whenever the room is occupied and the lights are off, in the next state, the motion sensors will activate, and the emergency lights will turn on.

- $G(p\land q\to r) \& G(r\to X(s))$ \\
  This formula breaks down the behavior into two separate conditions. Globally, when the room is occupied and the lights are off, the motion sensors activate. Once motion sensors are activated, the emergency lights will turn on in the next state.

- $G(p\land q\to X(r)) \& G(r\to X(s))$ \\
  Here, it states that globally, when the room is occupied and lights are off, in the next state, the motion sensors activate. And when motion sensors are activated, in the next state, emergency lights will turn on.

- $G(p\to(q\to(r\land X(s))))$ \\
  This formula suggests a cascading behavior, i.e., globally, whenever the room is occupied, and then the lights are off, the motion sensors will activate, and the emergency lights will turn on in the next time frame.

3. Verification:
The provided traces and the English description align with the following LTL formula:

$G((p\land q)\to(r\land X(s)))$.
This formula ensures that always, whenever the room is occupied (p) and the lights are off (q), the motion sensors will be activated (r), and in the subsequent time frame, the emergency lights will turn on (s).
\end{minipage}
}
\end{center}

\begin{center}
\fbox{
\begin{minipage}{1\textwidth}
\small
\textbf{Training: Example 5 - Go through each step}

Input Traces where “,” is the label separator, “;” is the state separator, and “\#” is the end of trace:

\texttt{!p,!q,r,s,t,u;p,!q,!r,s,!t,!u;!p,q,r,s,!t,u;1;1;1;1;1;1;1\#}

\texttt{!p,!q,r,s,t,!u;p,!q,r,s,!t,u;1;1;1;1;1;1;1;1\#}

\texttt{!p,q,r,s,!t,u;p,q,r,s,t,!u;p,q,r,s,!t,u;p,q,r,s,!t,!u;p,q,r,!s,!t,!u;\\!p,q,r,s,t,u;!p,!q,r,s,t,!u;p,!q,r,s,t,!u;!p,!q,r,!s,t,u;p,!q,r,s,t,!u\#}

\texttt{!p,!q,r,s,!t,!u;!p,!q,!r,!s,!t,u;1;1;1;1;1;1;1;1\#}

\texttt{!p,q,r,!s,t,u;!p,q,r,s,!t,u;!p,!q,r,!s,!t,!u;p,q,r,s,!t,!u;p,q,r,!s,t,u;\\!p,!q,r,!s,!t,!u;!p,!q,r,s,!t,!u;p,q,r,!s,t,!u;p,!q,r,s,!t,!u;p,q,r,s,t,u\#}

\texttt{p,!q,r,s,!t,!u;!p,q,!r,s,!t,u;!p,!q,r,s,t,u;1;1;1;1;1;1;1\#}

\texttt{p,q,!r,s,t,u;!p,q,!r,s,t,!u;1;1;1;1;1;1;1;1\#}

\texttt{p,!q,!r,!s,t,!u;!p,q,r,!s,t,u;!p,!q,r,s,!t,!u;p,q,r,!s,t,u;!p,q,r,!s,!t,!u;\\p,!q,r,s,t,u;!p,!q,r,!s,!t,!u;p,q,r,s,t,u;p,!q,r,s,t,!u;\\p,q,r,!s,t,!u;p,q,r,s,!t,u\#}

\texttt{p,q,r,s,!t,!u;!p,q,r,!s,t,!u;!p,q,r,s,t,u;p,!q,r,s,!t,!u;!p,q,r,s,t,!u;\\p,q,r,s,!t,u;p,!q,r,s,!t,!u;p,q,r,!s,t,!u;p,q,r,s,!t,!u;p,q,r,s,t,u\#}

\texttt{p,q,r,!s,t,u;!p,q,!r,s,t,u;p,!q,r,!s,t,u;1;1;1;1;1;1;1\#}

\textbf{English Description:} Either the system is in standby (p) or it's active (q) and always from then on processes tasks (r). This continues until it encounters an error (s) and in the next time period either sends an alert (t) or eventually restarts (u).

Most relevant formula: $(p|(q\&G(r)))U(s\&X(t|F(u)))$
\end{minipage}
}
\end{center}

\begin{center}
\fbox{
\begin{minipage}{\textwidth}
\small

ChatGPT: Alright, let's break down this problem step-by-step:

1. Observe and find patterns in traces:
From the traces, we can identify:

- Sequences where the system is active q and is consistently processing tasks r until it encounters an error s.
- Sequences where upon encountering an error s, in the next time period, an alert is sent t or eventually, the system restarts u.
- Sequences where the system is on standby p.

2. Generate 5-LTL formula:
Given the English Description: Either the system is in standby (p) or it's active (q) and always from then on processes tasks (r). This continues until an error is encountered (s) and in the next time period either sends an alert (t) or eventually restarts (u).

We can generate the following LTL formulas:

- $(p|(q\&G(r)))U(s\&X(t|F(u)))$: This formula captures the entire behavior succinctly. It states that the system is either on standby or when it's active, it will keep processing tasks. This goes on until an error is encountered. Upon encountering an error, in the next state, it will send an alert or at some point, the system restarts.

- $(p|G(q\&r))U(s\&X(t|F(u)))$: This formula says that the system is either on standby, or it is continuously active and processing tasks until an error occurs. When the error occurs, it will send an alert in the next state or eventually restart.

- $(p|G(q\to r))U(s\to(X(t)|F(u)))$: This formula suggests that the system is either on standby or when active, it processes tasks. When it encounters an error, it will either send an alert in the next state or at some point in the future, restart.

- $(p|G(q\&r))\&G(s\to(X(t)|F(u)))$: Breaking the formula into two global conditions. One handles the behavior of the system being on standby or active and processing tasks, and the other handles the behavior after encountering an error.

- $(p|G(q\&r))\&(s\to X(t|u))$: This formula suggests that if the system is on standby or is active and processing tasks, and if an error is encountered, in the next state, it will either send an alert or restart.

3. Verification:
Based on the traces provided and the English description, the most fitting formula seems to be:

$(p|(q\&G(r)))U(s\&X(t|F(u)))$.
This formula encapsulates the entire behavior. The system is either on standby or when active, always processes tasks. This behavior continues until an error is encountered, after which in the next state, an alert is sent or at some point, the system restarts.

\end{minipage}
}
\end{center}

\begin{center}
\fbox{
\begin{minipage}{1\textwidth}
\small
\textbf{Training: Example 6 - Go through each step}

Input Traces where “,” is the label separator, “;” is the state separator, and “\#” is the end of trace:

\texttt{p,!q,r;!p,q,r;!p,!q,!r;!p,q,!r;p,!q,r;!p,q,!r;p,q,r;!p,q,!r;p,!q,!r;!p,q,!r;\\!p,!q,!r;!p,q,!r;p,!q,r;!p,q,r;p,!q,r;p,q,!r;!p,!q,!r;!p,q,r;!p,q,r;p,q,!r\#}

\texttt{!p,q,!r;!p,q,!r;!p,q,!r;p,!q,!r;p,q,!r;p,!q,r;p,!q,!r;p,!q,!r;!p,q,r;\\!p,q,r;p,!q,r;p,!q,!r;p,!q,!r;p,!q,r;!p,q,r;\\!p,!q,!r;p,!q,r;p,!q,r;p,!q,!r;p,!q,!r;p,q,r\#}

\texttt{p,!q,r;p,!q,!r;!p,q,!r;p,q,r;p,!q,r;!p,q,r;p,!q,!r;p,!q,r;p,!q,!r;!p,q,!r;\\p,q,!r;!p,q,!r;!p,!q,!r;p,!q,r;p,!q,!r;p,!q,r;p,!q,r;p,!q,!r;p,!q,!r;p,q,r\#}

\texttt{!p,q,!r;!p,!q,!r;p,!q,r;!p,q,r;!p,q,!r;!p,q,!r;p,!q,r;p,q,r;p,q,!r;p,q,r;\\!p,q,r;!p,!q,!r;p,!q,r;p,!q,r;!p,q,r;p,!q,r;p,!q,r;p,q,r;p,q,!r;\\p,q,!r;p,!q,r;!p,q,r;p,!q,r;!p,q,!r;p,!q,r;!p,q,!r;p,!q,r;p,!q,r;\\p,!q,r;p,q,r\#}

\texttt{p,!q,r;!p,q,r;p,!q,r;p,!q,r;!p,q,!r;p,!q,r;p,!q,r;p,q,r;p,q,!r;!p,!q,!r;\\!p,q,!r;p,!q,r;!p,q,r;p,!q,!r;p,!q,r;p,!q,r;p,q,r;\\p,!q,r;!p,q,!r;!p,q,r;p,!q,r;p,q,r;p,!q,r;p,q,!r;!p,q,!r;\\p,!q,r;p,q,r;p,q,!r\#}

\texttt{!p,q,!r;p,!q,!r;!p,q,!r;p,q,!r;!p,!q,!r;p,!q,r;p,!q,r;p,!q,r;p,!q,!r;!p,q,r;\\p,q,!r;!p,q,!r;p,!q,!r;p,q,r;p,!q,r;p,!q,r;p,q,r;!p,q,!r;!p,q,r;p,!q,r;\\p,!q,!r;p,!q,r;!p,q,r;p,q,!r;!p,q,!r;!p,q,r;p,q,r;p,q,!r;!p,q,r;p,q,!r;\\!p,q,!r;p,q,r;p,q,!r\#}

\texttt{p,!q,!r;p,!q,!r;!p,q,!r;p,!q,!r;!p,!q,!r;!p,q,r;!p,q,!r;p,!q,r;!p,q,!r;\\!p,!q,!r;!p,!q,!r;p,!q,!r;p,!q,r;p,!q,r;!p,q,!r;p,!q,r;p,!q,r;p,!q,r;\\p,!q,!r;p,q,r\#}

\textbf{English Description:} "Whenever a request is submitted (r), it will continuously be in the processing state (p) until it is eventually accepted (q)."

Most relevant formula: $G(r \to (p U q))$
\end{minipage}
}
\end{center}

\begin{center}
\fbox{
\begin{minipage}{1\textwidth}
\small

\textbf{ChatGPT:} Alright, let's process the information given:

1. Observe and find patterns in traces:
From the traces, we can make out the following:

- There are sequences where after a request is submitted (r), it is in the processing state (p) for a while and eventually gets accepted (q).
- There are sequences where a request is submitted but it does not go into the processing state or get accepted.

2. Generate 5-LTL formula:
Given the English Description: "Whenever a request is submitted (r), it will continuously be in the processing state (p) until it is eventually accepted (q)."

We can generate the following LTL formulas:

- $G(r \to (p U q))$: This is a straightforward representation of the description. Whenever a request is submitted, it remains in the processing state until it gets accepted.

- $G(r \to X(p U q))$: This formula is slightly different from the first. It says that when a request is submitted, in the next state, it will enter the processing state and eventually get accepted.

- $G(r \& F(p) \& F(q))$: This formula suggests that whenever a request is made, it will eventually be processed and accepted but not necessarily in direct succession.

- $G(r \to (G(p) \; | \; F(q)))$: This formula states that once a request is made, it will either be continuously processed or will eventually be accepted.

- $G((r \& p) \to F(q))$: This formula means that if a request is made and it's in the processing state, it will eventually be accepted.

3. Verification:
To verify the most fitting formula, we will refer to the traces. From a preliminary look, many traces seem to fit the pattern of once a request (r) is made, it stays in the processing state (p) until it gets accepted (q).

The most fitting formula that describes the English description as well as adheres to the traces provided is:

$G(r \to (p U q))$.

This formula captures the behavior that whenever a request is submitted, it remains in the processing state until it gets accepted, which is in line with the given English description and the provided traces.
\end{minipage}
}
\end{center}

\end{document}

%% file: formula-list.tex
\renewcommand{\arraystretch}{2}
\begin{tabular}{{|c|>{\centering}p{2.1cm}|p{8cm}|c|}}
\hline
Index & Original formula & Input Description & Source \\
\hline
1 & $G((p  \limplies  F(q)) \land (r \limplies G(s)))$ &  In a financial monitoring system, it is always the case that when unusual account activity is detected (p), a fraud investigation is eventually initiated (q), and if regulatory compliance checks are required (r), continuous monitoring is implemented (s). For example, detecting suspicious transactions triggers a future fraud investigation, and if regulations are involved, ongoing compliance monitoring is maintained. & spot \\
\hline
2 & $F(p  \land  (q  \limplies  (r  \land  G(s))))$ & At some point, the system will enter a critical
 state and if a user tries to access it, then 
the system will deny access and 
consistently send alerts.
(system will enter a critical state=p,  
  user tries to access = q,  the system will 
deny access=r, 
send alerts=s) & spot \\
\hline
3 & $G((a \lor b) \limplies X((c)U(d \land F(e))))$ & In a smart home security system, 
it is always the case that if either a window (a)
 or a door (b) is opened, then in the 
subsequent moment, the system continuously
 checks for motion detection (c)
 until it identifies a potential intruder
 (d) and eventually activates an alarm (e). & spot \\
 \hline
4 & $G((a \lor b) \limplies X((c \land X(d))U(e)))$ & In a complex manufacturing process, it is
 always the case that when either a raw 
material (a) or a semi-finished product 
(b) is processed, then in the next phase,
 the system ensures a specific manufacturing
 step (c) is completed and immediately followed 
by another step (d), continuously until the final 
inspection (e) is reached.
 & spot \\
 \hline
5 & $G(p \limplies X((q)U(r \lor s)))$ & It is always the case that when a user logs in, they will remain logged in until they either log out or their session expires. (p = user logs in, q = user remained logs in , r = session expires, s = logged out) & spot \\
\hline
\end{tabular}

%% file: semantics-results-trace-description-sem1.tex
\begin{tabular}{{|c|>{\centering}p{2.5cm}|c|c|c|>{\centering}p{2.5cm}|c|c|c|c|}}
\hline
Formula & LLM & SAT & FIT & Manual & $\ourtool$ & SAT & FIT & Manual & Runtime \\
\hline
1 & $G(p  \limplies  X(F(q)  \land  (r  \limplies  X(G(s)))))$ & \checkmark & 8.99 & 3.75 & $G((p  \limplies  G((F(q)  \land  (r  \limplies  X(G(s)))))))$ & \checkmark & 36.11 & 4.75 & 47.58 \\
\hline
2 & $F(p)  \land  (q  \limplies  X(r  \land  F(s)))$ & \texttimes & 0.21 & 3.75 & $(F(p)  \land  (q  \limplies  X((r  \limplies  F(s)))))$ & \checkmark & 0.29 & 4.50 & 2889.43 \\
\hline
3 & $G((a  \lor  b)  \limplies  X(G(c) U (d  \land  F(e))))$ & \texttimes & 1.72 & 3.75 & $G(((a  \lor  b)  \limplies  F((G(c)  \lor  (d  \land  F(e))))))$ & \checkmark & 4.97 & 2.50 & 23.28 \\
\hline
4 & $G((a  \lor  b)  \limplies  F(c  \land  (X(d) U e)))$ & \checkmark & 2.32 & 3.50 & $G(((a U b)  \limplies  F((c  \land  (X(d) U e)))))$ & \checkmark & 2.53 & 3.50 & 226.99 \\
\hline
5 & $F(p  \limplies  X(q U (r  \lor  s)))$ & \checkmark & 0.60 & 0.75 & $G((p  \limplies  G((q U (r  \lor  s)))))$ & \checkmark & 8.79 & 4.25 & 64.37 \\
\hline
\end{tabular}

%% file: semantics-results-trace-description-sem2.tex
\begin{tabular}{{|c|>{\centering}p{2.5cm}|c|c|c|>{\centering}p{2.5cm}|c|c|c|c|}}
\hline
Formula & LLM & SAT & FIT & Manual & $\ourtool$ & SAT & FIT & Manual & Runtime \\
\hline
1 & $G(p  \limplies  (F(q)  \land  (r  \limplies  G(s))))$ & \checkmark & 0.37 & 3.75 & $G((p  \limplies  (F(q)  \land  (s  \lor  G( \lnot r)))))$ & \texttimes & 0.59 & 3.00 & 129.85 \\
\hline
2 & $F(p  \land  q  \limplies  X(r  \land  F(s)))$ & \checkmark & 0.77 & 4.25 & $F(((p  \land  q)  \limplies  X((r  \land  F(s)))))$ & \checkmark & 0.61 & 4.25 & 45.35 \\
\hline
3 & $G((a  \lor  b)  \limplies  X(G(c) U (d  \land  F(e))))$ & \texttimes & 0.43 & 3.75 & $G(((a  \lor  b)  \limplies  F((G(c)  \limplies  (d  \land  F(e))))))$ & \checkmark & 0.54 & 5.00 & 125.89 \\
\hline
4 & $G(G((a  \lor  b)  \limplies  X(c  \land  Xd)) U e)$ & \texttimes & 0.73 & 4.25 & $G((G(((a  \lor  b)  \limplies  X((c  \land  X(d))))) U e))$ & \texttimes & 0.27 & 4.25 & 4.60 \\
\hline
5 & $GF(p  \limplies  X(q U (r  \lor  s)))$ & \checkmark & 0.57 & 1.00 & $G(F((p  \limplies  F((q U (r  \lor  s))))))$ & \checkmark & 0.69 & 4.25 & 39.92 \\
\hline
\end{tabular}

%% file: semantics-results-description.tex
\begin{tabular}{{|c|>{\centering}p{3cm}|c|c|c|>{\centering}p{3cm}|c|c|c|}}
\hline
Formula & LLM with traces & SAT & FIT & Manual & LLM without traces & SAT & FIT & Manual \\
\hline
1 & $G(p  \limplies  X(F(q)  \land  (r  \limplies  X(G(s)))))$ & \checkmark & 8.99 & 3.75 & $G(p  \land  r) \limplies (F(q)  \land  (r  \land(G(s))))$ &  \checkmark & 0.10  & 3 \\
\hline
2 &$F(p  \land  q  \limplies  X(r  \land  F(s)))$ & \checkmark & 0.77 & 4.25& $Fp  \land G(q  \limplies  X(r \land (G(s))))$ &  \texttimes & 0.10  & 3.5 \\
\hline
3 &$G((a  \lor  b)  \limplies  F((c) U (d  \land  F(e))))$ & \texttimes & 3.51 & 5 & $G((a  \lor  b)  \limplies  X(G(c) U (d  \land  X(e))))$  &  \texttimes & -0.009 & 3.5 \\
\hline
4 & $G(G((a  \lor  b)  \limplies  X(c  \land  Xd)) U e)$ & \texttimes & 0.27 & 4.25& $G((a  \lor  b)  \limplies  X(c  \land  d) U e)$ &  \texttimes & 0.13 & 4.25 \\
\hline
5 &$G(p  \limplies  (q U (r  \lor  s)))$ & \checkmark & 4.36 & 3.00& $G (p \limplies X(q U (s \lor  (r \land \lnot q))))$  & \checkmark & 0.53  & 1.75 \\
\hline
\end{tabular}